\documentclass{article}

\PassOptionsToPackage{numbers, compress, sectionbib}{natbib}
\usepackage[final]{neurips_2021}

\usepackage{chapterbib}
\usepackage[utf8]{inputenc} %
\usepackage[T1]{fontenc}    %
\usepackage[colorlinks=true,citecolor=black,urlcolor=black,linkcolor=black]{hyperref}       %
\usepackage{url}            %
\usepackage{booktabs}       %
\usepackage{amsfonts}       %
\usepackage{nicefrac}       %
\usepackage{microtype}      %
\usepackage{xcolor}         %

\usepackage{amsmath}
\usepackage{graphicx}
\usepackage{algorithm,algorithmic}

\usepackage{comment}
\usepackage{cancel}

\usepackage{subcaption}
\usepackage{amsmath}
\usepackage{amsfonts}
\usepackage{nicefrac}
\usepackage{xfrac}
\usepackage[capitalize,nameinlink]{cleveref}
\usepackage{booktabs,tabularx,array}
\usepackage[usestackEOL]{stackengine} %
\usepackage{multicol} %
\usepackage[normalem]{ulem}
\usepackage{enumerate}
\usepackage{wrapfig}

\usepackage{mathtools}
\usepackage{xspace}
\usepackage{amssymb}

\makeatletter
\def\smalloverbrace#1{\mathop{\vbox{\m@th\ialign{##\crcr\noalign{\kern3\p@}%
  \tiny\downbracefill\crcr\noalign{\kern3\p@\nointerlineskip}%
  $\hfil\displaystyle{#1}\hfil$\crcr}}}\limits}
\makeatother

\newcommand{\inv}{^{-1}}
\newcommand{\ExpSymb}{\mathbb{E}}
\newcommand{\Exp}[2]{\ExpSymb_{#1}\left[#2\right]}

\DeclarePairedDelimiterX{\dbar}[2]{[}{]}{#1\,\delimsize\|\,#2}
\newcommand{\KL}[2]{\mathrm{D}_\textrm{KL} \dbar*{#1}{#2}}

\DeclareMathOperator{\diag}{diag}

\crefname{section}{Sec.}{Secs.}
\crefname{flatappendix}{App.}{Apps.}
\crefname{algorithm}{Alg.}{Algs.}

\usepackage{tikz,pgfplots}
\usetikzlibrary{plotmarks,shapes,shapes.arrows,positioning,calc,spy,intersections}
\pgfplotsset{compat=newest} 
\pgfkeys{/pgf/number format/.cd,1000 sep={}}
\usepackage{placeins}

\newlength\figureheight
\newlength\figurewidth

\pgfplotsset{every axis/.append style={
  grid style={line width=0.6pt,dotted,gray}}}

\pgfplotsset{every axis/.append style={
		legend style={inner xsep=1pt, inner ysep=0.5pt, nodes={inner sep=1pt, text depth=0.1em},draw=none,fill=none}
}}    
\definecolor{mycolor0}{rgb}{0.2667,0.4471,0.7098}
\definecolor{mycolor1}{rgb}{0.1647,0.6706,0.3804}
\definecolor{mycolor2}{rgb}{0.8275,0.2627,0.3059}
\definecolor{mycolor3}{rgb}{0.5216,0.4392,0.7176}
\definecolor{mycolor4}{rgb}{0.8118,0.7255,0.4118}
\definecolor{mycolor5}{rgb}{0.2745,0.7176,0.8157}

\definecolor{mylcolor0}{rgb}{0.6902,0.7686,0.8863}
\definecolor{mylcolor1}{rgb}{0.5451,0.8902,0.6941}
\definecolor{mylcolor2}{rgb}{0.9412,0.7490,0.7647}
\definecolor{mylcolor3}{rgb}{0.8627,0.8392,0.9176}
\definecolor{mylcolor4}{rgb}{0.9569,0.9373,0.8667}
\definecolor{mylcolor5}{rgb}{0.7529,0.9020,0.9373}
\definecolor{mylcolor6}{rgb}{0.8750,0.8750,0.8750}

\usepackage{bm}
\newcommand{\mathbold}[1]{\bm{#1}}
\newcommand{\mbf}[1]{\mathbf{#1}}

\usepackage{xspace}
\newcommand{\eg}{\textit{e.g.}\xspace}
\newcommand{\ie}{\textit{i.e.}\xspace}
\newcommand{\cf}{\textit{cf.}\xspace}

\newcommand{\R}{\mathbb{R}}    %
\newcommand{\N}{\mathrm{N}}  
\DeclareMathOperator{\tr}{tr}

\DeclareMathOperator{\argmin}{arg\,min}
\DeclareMathOperator{\argmax}{arg\,max}

\newcommand{\GP}{\mathcal{GP}}
\newcommand{\ELBO}{\mathcal{L}}

\newcommand{\cL}{\mathcal{L}}

\newcommand{\order}{\mathcal{O}}

\newcommand{\valpha}[0]{\mathbold{\alpha}}
\newcommand{\vbeta}[0]{\mathbold{\beta}}

\newcommand{\veta}[0]{\mathbold{\eta}}
\newcommand{\vmu}[0]{\mathbold{\mu}}

\newcommand{\vxi}[0]{\mathbold{\xi}}
\newcommand{\vtheta}[0]{\mathbold{\theta}}

\newcommand{\MLambda}[0]{\mathbold{\Lambda}}

\renewcommand{\mid}[0]{\,|\,}

\newcommand{\vlambda}[0]{\mathbold{\lambda}}

\newcommand{\data}{\mathcal{D}}

\newcommand{\indu}{\vu}

\newcommand{\myexpect}{\mathbb{E}}

\newcommand{\dee}{\mathrm{d}}
\newcommand{\mint}{\textstyle\int}

\newcommand{\va}{\mbf{a}}

\newcommand{\vf}{\mbf{f}}
\newcommand{\vg}{\mbf{g}}

\newcommand{\vk}{\mbf{k}}

\newcommand{\vm}{\mbf{m}}

\newcommand{\vu}{\mbf{u}}
\newcommand{\vv}{\mbf{v}}

\newcommand{\vx}{\mbf{x}}
\newcommand{\vy}{\mbf{y}}
\newcommand{\vz}{\mbf{z}}

\newcommand{\MF}{\mbf{F}}

\newcommand{\MI}{\mbf{I}}

\newcommand{\MK}{\mbf{K}}
\newcommand{\ML}{\mbf{L}}

\newcommand{\MR}{\mbf{R}}
\newcommand{\MS}{\mbf{S}}
\newcommand{\MT}{\mbf{T}}

\newcommand{\MX}{\mbf{X}}

\newcommand{\MZ}{\mbf{Z}}

\newcommand{\MKff}{\mbf{K}_{\mbf{f}\mbf{f}}}
\newcommand{\MKuu}{\mbf{K}_{\mbf{u}\mbf{u}}}
\newcommand{\MKsu}{\mbf{K}_{\star\mbf{u}}}
\newcommand{\MKss}{\mbf{K}_{\star\star}}
\newcommand{\MKus}{\mbf{K}_{\mbf{u}\star}}

\newcommand{\mprod}{\textstyle\prod}
\newcommand{\msum}{\textstyle\sum}

\newcommand{\rnd}[1]{\left(#1\right)}
\newcommand{\sqr}[1]{\left[#1\right]}

\definecolor{mygreen}{rgb}{0.09, 0.45, 0.27}

\renewcommand{\paragraph}[1]{\textbf{#1}~~}

\newcommand{\nipstitle}[1]{{%
    \def\toptitlebar{\hrule height4pt \vskip .25in \vskip -\parskip} 
    \def\bottomtitlebar{\vskip .29in \vskip -\parskip \hrule height1pt \vskip .09in} 
    \phantomsection\hsize\textwidth\linewidth\hsize%
    \vskip 0.1in%
    \toptitlebar%
    \begin{minipage}{\textwidth}%
        \centering{\LARGE\bf #1\par}%
    \end{minipage}%
    \bottomtitlebar%
    \addcontentsline{toc}{section}{#1}%
}}

\title{Dual Parameterization of Sparse Variational Gaussian Processes}

\author{%
  Vincent Adam$^*$\\
  Aalto University / Secondmind.ai\\
  Espoo, Finland / Cambridge, UK\\
  \texttt{vincent.adam@aalto.fi} \\[-1em]
  \hphantom{Aalto University / Secondmind.ai}
  \And
  Paul E.\ Chang\thanks{Both authors contributed equally.}\\
  Aalto University\\
  Espoo, Finland\\
  \texttt{paul.chang@aalto.fi} \\[-1em]
  \hphantom{Aalto University / Secondmind.ai}
  \AND
  Mohammad Emtiyaz Khan \\
  RIKEN Center for AI Project\\
  Tokyo, Japan\\
  \texttt{emtiyaz.khan@riken.jp} \\[-1em]
  \hphantom{Aalto University / Secondmind.ai}
  \And
  Arno Solin \\
  Aalto University\\
  Espoo, Finland\\
  \texttt{arno.solin@aalto.fi} \\[-1em]
  \hphantom{Aalto University / Secondmind.ai}
}

\begin{document}

\maketitle

\begin{abstract}
Sparse variational Gaussian process (SVGP) methods are a common choice for non-conjugate Gaussian process inference because of their computational benefits. In this paper, we improve their computational efficiency by using a dual parameterization where each data example is assigned dual parameters, similarly to site parameters used in expectation propagation. Our dual parameterization speeds-up inference using natural gradient descent, and provides a tighter evidence lower bound for hyperparameter learning. The approach has the same memory cost as the current SVGP methods, but it is faster and more accurate.

\end{abstract}

\begingroup
\let\clearpage\relax

\section{Introduction}
Gaussian processes (GPs, \cite{rasmussen2006gaussian}) have become ubiquitous models in the probabilistic machine learning toolbox, but their application is challenging due to two issues: poor $\order(n^3)$ scaling in the number of data points, $n$, and challenging approximate inference in non-conjugate (non-Gaussian) models. In recent years, variational inference has become the go-to solution to overcome these problems, where sparse variational GP methods \cite{titsias2009variational} tackle both the non-conjugacy and high computation cost. The computation is reduced to $\order(nm^2)$ by using a small number of $m \ll n$ inducing-input locations. For large problems, the SVGP framework is a popular choice as it enables fast stochastic training, and reduces the cost 
to $\order(m^3 + n_\mathrm{b} m^2)$ per step, where $n_\mathrm{b}$ is the batch size \cite{hensman2013gaussian,Hensman+Matthews+Ghahramani:2015}.

It is a common practice in SVGP to utilize the standard mean-covariance parameterization which requires $\order(m^2)$ memory. Inference is carried out by optimizing an objective $\ELBO(q)$ that uses a Gaussian distribution $q$ parameterized by parameters $\vxi= (\vm, \ML)$ where $\vm$ is the mean and $\ML$ is the Cholesky factor of the covariance matrix. We will refer to the SVGP methods using such parameterization as the $q$-SVGP methods. The advantage of this formulation is that, for log-concave likelihoods, 
the objective is convex and gradient-based optimization works well \cite{pmlr-v15-challis11a}. The optimization can further be improved by using natural gradient descent (NGD) which is shown to be less sensitive to learning rates \cite{salimbeni2018natural}. The NGD algorithm with $q$-SVGP parameterization is currently the state-of-the-art and available in the existing software implementations such as GPflow~\cite{GPflow:2017} and GPyTorch~\cite{GPyTorch}.

An alternate parameterization to the $q$-SVGP parameterization is the one where every likelihood is assigned two sets of parameters which require $\order(n)$ memory. Existence of such parameterizations was initially shown by \citet{csato2002sparse} for general GP models, and later on extended to variational objectives for GPs \cite{nickisch2008approximations, opper2009variational}, and also to latent Gaussian models by using Lagrangian duality \cite{khan2013fast, khan2014decoupled}.
Due to this later connection, we refer to this parameterization as the \emph{dual parameterization} where the parameters are the Lagrange multipliers, ensuring that the marginal mean and variance of each latent function is consistent to the marginals obtained by the full GP~\citep{khan2014decoupled}.
Expectation propagation (EP, \cite{minka2001expectation}) too naturally employ such parameterizations, but by using site parameters. Although unrelated to duality, such methods are popular for GP inference, and due to this connection, we will refer to the methods using dual parameterization as $t$-SVGP (the letter `$t$' refers to the sites).
To the best of our knowledge, the dual parameterization for SVGP has only been used to speed up computation and inference for the specific case of Markovian GPs \cite{chang2020fast,wilkinson2021sparse}.

The main contribution of this work is to introduce the dual parameterization for SVGP and show that it speeds up both the learning and inference. For inference, we show that the dual parameters are automatically obtained through a different formulation of NGD, written in terms of the expectation parameters \cite{khan2017conjugate, khan2018fast, khan2018}. The formulation is fast since it avoids the use of sluggish automatic differentiation to compute the natural gradients. We also match the typical $\order(m^2)$ memory complexity in other SVGP methods by introducing a \emph{tied} parametrization.
For learning, we show that the dual parameterization results in a \emph{tighter} lower bound to the marginal likelihood, which speed-up the hyperparameter optimization (see \cref{fig:bound}). 
We provide extensive evaluation on benchmark data sets, which confirms our findings. Our work attempts to revive the dual parameterization, which was popular in the early 2000s, but was somehow forgotten and not used in the recent SVGP algorithms.
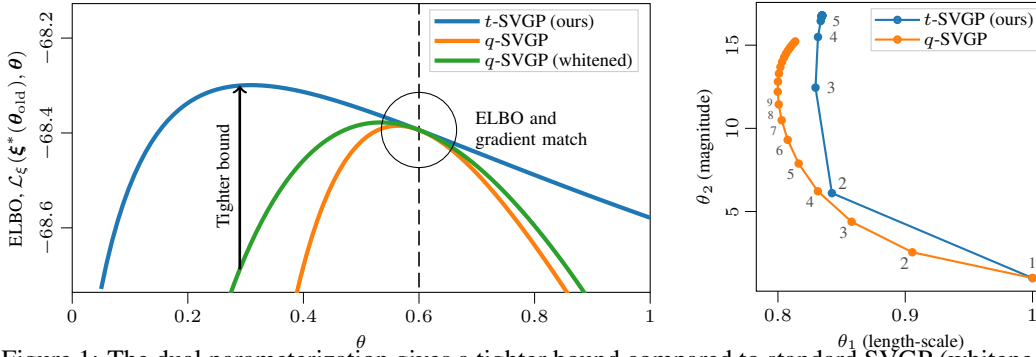
\begin{figure}[t]
  \centering\scriptsize
  \setlength{\figurewidth}{.55\textwidth}
  \setlength{\figureheight}{.5\figurewidth}
  \pgfplotsset{scale only axis,y tick label style={rotate=90}}
  \begin{subfigure}[t]{.65\textwidth}
\begin{tikzpicture}

\definecolor{color0}{rgb}{0.12156862745098,0.466666666666667,0.705882352941177}
\definecolor{color1}{rgb}{1,0.498039215686275,0.0549019607843137}
\definecolor{color2}{rgb}{0.172549019607843,0.627450980392157,0.172549019607843}

\begin{axis}[
height=\figureheight,
legend cell align={left},
legend style={fill opacity=0.8, draw opacity=1, text opacity=1, draw=white!80!black},
tick align=outside,
tick pos=left,
width=\figurewidth,
x grid style={white!69.0196078431373!black},
xlabel={$\theta\vphantom{\theta_1}$},
xmin=0, xmax=1,
xtick style={color=black},
y grid style={white!69.0196078431373!black},
ylabel={ELBO, $\cL_{\xi}(\vxi^*(\vtheta_\mathrm{old}), \vtheta)$},
ymin=-68.7359123320244, ymax=-68.1273432512982,
ytick style={color=black}
]
\path [draw=black, semithick, dash pattern=on 5.55pt off 2.4pt]
(axis cs:0.6,-78.7287481922727)
--(axis cs:0.6,-58.2991732344337);

\addplot [ultra thick, color0]
table {%
0.05 -68.7287481922727
0.0595959595959596 -68.6739495844939
0.0691919191919192 -68.6264791880295
0.0787878787878788 -68.5850109970875
0.0883838383838384 -68.5485800198754
0.097979797979798 -68.5164503830629
0.107575757575758 -68.4880408586256
0.117171717171717 -68.4628800220513
0.126767676767677 -68.4405776944248
0.136363636363636 -68.4208058454304
0.145959595959596 -68.4032852735549
0.155555555555556 -68.3877759751593
0.165151515151515 -68.3740699650531
0.174747474747475 -68.3619857853023
0.184343434343434 -68.3513642137888
0.193939393939394 -68.3420648492258
0.203535353535354 -68.333963352016
0.213131313131313 -68.3269491861969
0.222727272727273 -68.3209237511851
0.232323232323232 -68.3157988214982
0.241919191919192 -68.3114952330996
0.251515151515152 -68.3079417695482
0.261111111111111 -68.3050742116777
0.270707070707071 -68.302834522308
0.28030303030303 -68.3011701433363
0.28989898989899 -68.3000333870066
0.299494949494949 -68.2993809065942
0.309090909090909 -68.2991732344337
0.318686868686869 -68.2993743773368
0.328282828282828 -68.2999514611485
0.337878787878788 -68.3008744175494
0.347474747474747 -68.3021157073215
0.357070707070707 -68.3036500751978
0.366666666666667 -68.305454332162
0.376262626262626 -68.3075071616784
0.385858585858586 -68.3097889468443
0.395454545454545 -68.3122816158877
0.405050505050505 -68.3149685037879
0.414646464646465 -68.3178342281059
0.424242424242424 -68.3208645773628
0.433838383838384 -68.3240464105268
0.443434343434343 -68.3273675663526
0.453030303030303 -68.3308167814793
0.462626262626263 -68.334383616327
0.472222222222222 -68.3380583879516
0.481818181818182 -68.341832109119
0.491414141414141 -68.3456964329468
0.501010101010101 -68.3496436025393
0.510606060606061 -68.3536664051073
0.52020202020202 -68.3577581301221
0.52979797979798 -68.3619125311034
0.539393939393939 -68.3661237906871
0.548989898989899 -68.3703864886539
0.558585858585859 -68.3746955726381
0.568181818181818 -68.3790463312633
0.577777777777778 -68.3834343694793
0.587373737373737 -68.3878555858971
0.596969696969697 -68.3923061519415
0.606565656565657 -68.3967824926566
0.616161616161616 -68.4012812690179
0.625757575757576 -68.4057993616194
0.635353535353535 -68.4103338556141
0.644949494949495 -68.4148820268025
0.654545454545455 -68.419441328769
0.664141414141414 -68.4240093809806
0.673737373737374 -68.4285839577647
0.683333333333333 -68.4331629780964
0.692929292929293 -68.4377444961269
0.702525252525253 -68.4423266923938
0.712121212121212 -68.4469078656598
0.721717171717172 -68.4514864253276
0.731313131313131 -68.456060884388
0.740909090909091 -68.4606298528579
0.75050505050505 -68.4651920316718
0.76010101010101 -68.4697462069911
0.76969696969697 -68.4742912448998
0.779292929292929 -68.478826086458
0.788888888888889 -68.4833497430856
0.798484848484848 -68.4878612922531
0.808080808080808 -68.4923598734561
0.817676767676768 -68.4968446844524
0.827272727272727 -68.5013149777442
0.836868686868687 -68.5057700572871
0.846464646464646 -68.5102092754095
0.856060606060606 -68.514632029928
0.865656565656566 -68.5190377614454
0.875252525252525 -68.5234259508183
0.884848484848485 -68.5277961167816
0.894444444444444 -68.5321478137223
0.904040404040404 -68.5364806295881
0.913636363636364 -68.540794183926
0.923232323232323 -68.5450881260388
0.932828282828283 -68.5493621332548
0.942424242424242 -68.5536159092994
0.952020202020202 -68.5578491827663
0.961616161616162 -68.5620617056783
0.971212121212121 -68.5662532521341
0.980808080808081 -68.5704236170341
0.99040404040404 -68.574572614881
1 -68.5787000786511
};
\addlegendentry{$t$-SVGP (ours)}
\addplot [ultra thick, color1]
table {%
0.05 -101.682288853233
0.0595959595959596 -94.8929677500473
0.0691919191919192 -90.0407751240499
0.0787878787878788 -86.4171998026786
0.0883838383838384 -83.6204677069246
0.097979797979798 -81.4058939312263
0.107575757575758 -79.6160453184935
0.117171717171717 -78.1451054858967
0.126767676767677 -76.9193754636988
0.136363636363636 -75.8859853818891
0.145959595959596 -75.0060490029086
0.155555555555556 -74.2503475237155
0.165151515151515 -73.5965162755892
0.174747474747475 -73.0271577548262
0.184343434343434 -72.5285439893823
0.193939393939394 -72.0897043733023
0.203535353535354 -71.7017718595233
0.213131313131313 -71.3575061170017
0.222727272727273 -71.0509402789043
0.232323232323232 -70.7771155290235
0.241919191919192 -70.531879112773
0.251515151515152 -70.3117288087226
0.261111111111111 -70.1136918846784
0.270707070707071 -69.9352299600852
0.28030303030303 -69.7741635479937
0.28989898989899 -69.6286117001696
0.299494949494949 -69.4969433546299
0.309090909090909 -69.3777378303588
0.318686868686869 -69.2697525314332
0.328282828282828 -69.1718963770734
0.337878787878788 -69.0832078119944
0.347474747474747 -69.0028365058708
0.357070707070707 -68.930028042565
0.366666666666667 -68.8641110474024
0.376262626262626 -68.8044863132662
0.385858585858586 -68.7506175746306
0.395454545454545 -68.7020236468439
0.405050505050505 -68.6582717020349
0.414646464646465 -68.6189714954196
0.424242424242424 -68.5837703900186
0.433838383838384 -68.5523490545332
0.443434343434343 -68.5244177312488
0.453030303030303 -68.4997129883433
0.462626262626263 -68.477994885262
0.472222222222222 -68.4590444916381
0.481818181818182 -68.4426617097187
0.491414141414141 -68.4286633581674
0.501010101010101 -68.4168814816639
0.510606060606061 -68.4071618561083
0.52020202020202 -68.3993626637225
0.52979797979798 -68.3933533161769
0.539393939393939 -68.3890134069118
0.548989898989899 -68.3862317765976
0.558585858585859 -68.3849056778507
0.568181818181818 -68.3849400272362
0.577777777777778 -68.386246734196
0.587373737373737 -68.3887440979328
0.596969696969697 -68.3923562644322
0.606565656565657 -68.3970127368084
0.616161616161616 -68.402647933039
0.625757575757576 -68.4092007858737
0.635353535353535 -68.4166143803545
0.644949494949495 -68.4248356249356
0.654545454545455 -68.4338149526574
0.664141414141414 -68.44350604928
0.673737373737374 -68.4538656055769
0.683333333333333 -68.4648530914057
0.692929292929293 -68.476430549319
0.702525252525253 -68.4885624058693
0.712121212121212 -68.5012152988036
0.721717171717172 -68.5143579187317
0.731313131313131 -68.5279608637722
0.740909090909091 -68.5419965060923
0.75050505050505 -68.5564388691537
0.76010101010101 -68.5712635146921
0.76969696969697 -68.5864474386324
0.779292929292929 -68.6019689750499
0.788888888888889 -68.6178077075239
0.798484848484848 -68.6339443871882
0.808080808080808 -68.6503608570121
0.817676767676768 -68.6670399816211
0.827272727272727 -68.6839655823653
0.836868686868687 -68.7011223770177
0.846464646464646 -68.7184959238898
0.856060606060606 -68.7360725699015
0.865656565656566 -68.7538394023178
0.875252525252525 -68.771784203823
0.884848484848485 -68.7898954107778
0.894444444444444 -68.8081620742379
0.904040404040404 -68.8265738236974
0.913636363636364 -68.8451208332244
0.923232323232323 -68.8637937898939
0.932828282828283 -68.8825838641914
0.942424242424242 -68.901482682526
0.952020202020202 -68.9204823013468
0.961616161616162 -68.9395751830372
0.971212121212121 -68.9587541732915
0.980808080808081 -68.9780124799489
0.99040404040404 -68.9973436530939
1 -69.0167415664468
};
\addlegendentry{$q$-SVGP}
\addplot [ultra thick, color2]
table {%
0.05 -70.5985540953432
0.0595959595959596 -70.4182515949121
0.0691919191919192 -70.2590936285146
0.0787878787878788 -70.1167506571688
0.0883838383838384 -69.9881878925881
0.097979797979798 -69.8711812997005
0.107575757575758 -69.7640438106291
0.117171717171717 -69.6654599467688
0.126767676767677 -69.5743806189629
0.136363636363636 -69.4899533429375
0.145959595959596 -69.4114743131112
0.155555555555556 -69.338354512328
0.165151515151515 -69.2700951413617
0.174747474747475 -69.206269416893
0.184343434343434 -69.1465088309352
0.193939393939394 -69.0904926046303
0.203535353535354 -69.0379394736718
0.213131313131313 -68.9886012050439
0.222727272727273 -68.9422574192195
0.232323232323232 -68.8987114104292
0.241919191919192 -68.8577867396132
0.251515151515152 -68.819324432438
0.261111111111111 -68.7831806560851
0.270707070707071 -68.7492247785288
0.28030303030303 -68.7173377360846
0.28989898989899 -68.6874106514517
0.299494949494949 -68.6593436568367
0.309090909090909 -68.6330448861688
0.318686868686869 -68.6084296076437
0.328282828282828 -68.5854194734434
0.337878787878788 -68.5639418678574
0.347474747474747 -68.5439293384873
0.357070707070707 -68.525319097955
0.366666666666667 -68.5080525857232
0.376262626262626 -68.4920750814024
0.385858585858586 -68.4773353623416
0.395454545454545 -68.4637853994603
0.405050505050505 -68.4513800862318
0.414646464646465 -68.440076996508
0.424242424242424 -68.4298361675227
0.433838383838384 -68.4206199049477
0.443434343434343 -68.4123926073248
0.453030303030303 -68.4051206075718
0.462626262626263 -68.3987720295763
0.472222222222222 -68.3933166581608
0.481818181818182 -68.3887258209253
0.491414141414141 -68.3849722806691
0.501010101010101 -68.3820301372583
0.510606060606061 -68.3798747379431
0.52020202020202 -68.3784825952556
0.52979797979798 -68.3778313117169
0.539393939393939 -68.3778995106785
0.548989898989899 -68.3786667726975
0.558585858585859 -68.3801135769137
0.568181818181818 -68.3822212469569
0.577777777777778 -68.3849719009646
0.587373737373737 -68.3883484053341
0.596969696969697 -68.3923343318739
0.606565656565657 -68.3969139180544
0.616161616161616 -68.4020720300884
0.625757575757576 -68.4077941285984
0.635353535353535 -68.4140662366544
0.644949494949495 -68.4208749099835
0.654545454545455 -68.4282072091744
0.664141414141414 -68.4360506737165
0.673737373737374 -68.4443932977274
0.683333333333333 -68.4532235072368
0.692929292929293 -68.4625301389054
0.702525252525253 -68.4723024200726
0.712121212121212 -68.4825299500294
0.721717171717172 -68.4932026824281
0.731313131313131 -68.5043109087461
0.740909090909091 -68.5158452427249
0.75050505050505 -68.5277966057181
0.76010101010101 -68.5401562128829
0.76969696969697 -68.5529155601561
0.779292929292929 -68.5660664119618
0.788888888888889 -68.579600789601
0.798484848484848 -68.593510960277
0.808080808080808 -68.6077894267151
0.817676767676768 -68.6224289173371
0.827272727272727 -68.6374223769568
0.836868686868687 -68.6527629579589
0.846464646464646 -68.6684440119364
0.856060606060606 -68.6844590817541
0.865656565656566 -68.700801894012
0.875252525252525 -68.7174663518866
0.884848484848485 -68.7344465283249
0.894444444444444 -68.7517366595715
0.904040404040404 -68.7693311390085
0.913636363636364 -68.7872245112899
0.923232323232323 -68.8054114667536
0.932828282828283 -68.8238868360954
0.942424242424242 -68.8426455852904
0.952020202020202 -68.8616828107462
0.961616161616162 -68.8809937346775
0.971212121212121 -68.9005737006878
0.980808080808081 -68.9204181695491
0.99040404040404 -68.9405227151665
1 -68.9608830207195
};
\addlegendentry{$q$-SVGP (whitened)}

\coordinate (elbo) at (axis cs:0.6, -68.3937171114626);
\coordinate (cvi) at (axis cs:0.28989898989899, -68.3000333870066);
\coordinate (svgpw) at (axis cs:0.28989898989899, -68.6874106514517);

\draw[->,line width=1pt] (svgpw)--(cvi);

\node[circle,minimum size=1cm,draw=black] at (elbo) {};

\node[right of=elbo,xshift=0.5cm,text width=1.5cm] {ELBO and gradient match};
\node[right of=cvi,xshift=-1.2cm,yshift=-1.2cm,rotate=90] {Tighter bound};

\end{axis}

\end{tikzpicture}
   \end{subfigure}
  \hfill
  \begin{subfigure}[t]{.34\textwidth}
    \setlength{\figurewidth}{\figureheight}  
\begin{tikzpicture}

\definecolor{color0}{rgb}{0.12156862745098,0.466666666666667,0.705882352941177}
\definecolor{color1}{rgb}{1,0.498039215686275,0.0549019607843137}

\begin{axis}[
height=\figureheight,
legend cell align={left},
legend style={fill opacity=0.8, draw opacity=1, text opacity=1, draw=white!80!black},
tick align=outside,
tick pos=left,
width=\figurewidth,
x grid style={white!69.0196078431373!black},
xlabel={$\theta_1$ (length-scale)},
xmin=0.783, xmax=1.01000848615828,
xtick style={color=black},
y grid style={white!69.0196078431373!black},
ylabel={$\theta_2$ (magnitude)},
ymin=0.211003811543253, ymax=17.5689199575917,
ytick style={color=black}
]
\addplot [thick, color0, mark=*, mark size=1.2, mark options={solid}]
table {%
1 1
0.842399878983801 6.10605160924438
0.829506708317353 12.4466987770672
0.831448473265075 15.4931995483585
0.833621557410984 16.4503226029425
0.834443422500925 16.7081025135444
0.834135009820684 16.708139596861
0.834663582536103 16.7777027822339
0.834639129721135 16.7777065040897
0.834633795066711 16.7777100366635
0.83462745739182 16.7777500257968
0.834633892375011 16.777753764489
0.834625273441405 16.7778392674744
0.834634282038595 16.777842708944
0.8346097328308 16.7798963078209
0.834649351354983 16.7798983884209
0.834652719302545 16.7799052491452
0.834649716776734 16.7799112664722
0.834653135215288 16.779918885539
0.834650255612117 16.7799237691349
};
\addlegendentry{$t$-SVGP (ours)}
\addplot [thick, color1, mark=*, mark size=1.2, mark options={solid}]
table {%
1 1
0.905572909212576 2.55010924459064
0.857874728396063 4.38362034846977
0.831373873821824 6.21655977652568
0.816263509796623 7.8821258425781
0.80765829195396 9.30924004560519
0.802940596601701 10.4878335602194
0.800627427939646 11.4400394484114
0.799830276834489 12.200979791148
0.799996210905919 12.807018043569
0.800769171073399 13.2914590998882
0.801916395887349 13.6819139125991
0.803283894194426 14.0002868908664
0.804768448193289 14.2635756592169
0.806300847574512 14.4847194099851
0.807836242981008 14.6732862903108
0.809344275796063 14.8365771827757
0.810806201568407 14.9799081348002
0.812210337033262 15.1073092497192
0.81355011618286 15.2219985344229
};
\addlegendentry{$q$-SVGP}

\node[yshift=6pt,text=black!70] at (axis cs:1,1) {\tiny 1};
\node[yshift=4pt,xshift=4pt,text=black!70] at (axis cs:0.842399878983801,6.10605160924438) {\scalebox{0.95}{2}};
\node[yshift=0pt,xshift=6pt,text=black!70] at (axis cs:0.829506708317353,12.4466987770672) {\scalebox{0.9}{3}};
\node[yshift=0pt,xshift=6pt,text=black!70] at (axis cs:0.831448473265075,15.4931995483585) {\scalebox{0.85}{4}};
\node[yshift=0pt,xshift=6pt,text=black!70] at (axis cs:0.833621557410984,16.4503226029425) {\scalebox{0.8}{5}};

\foreach \xValue/\yValue/\z [count=\i from 2] in {%
  0.905572909212576/2.55010924459064/0.95,
  0.857874728396063/4.38362034846977/0.9,
  0.831373873821824/6.21655977652568/0.85,
  0.816263509796623/7.88212584257810/0.8,
  0.807658291953960/9.30924004560519/0.75,
  0.802940596601701/10.4878335602194/0.7,
  0.800627427939646/11.4400394484114/0.65,
  0.799830276834489/12.2009797911480/0.6} {
  \edef\temp{\noexpand\node[yshift=-4pt,xshift=-3pt,text=black!70] at (axis cs:\xValue,\yValue) {\scalebox{\z}{\i}};}
  \temp
}

\end{axis}

\end{tikzpicture}
   \end{subfigure}
  \vspace*{-1.5em}
  \caption{%
  The dual parameterization gives a tighter bound compared to standard SVGP (whitened/ unwhitened) as shown on the left for a GP classification tasks and varying kernel magnitude $\theta$. The tighter bound helps take longer steps, speeding up convergence of hyperparameters, as shown on the right for the {\em banana classification} task with coordinate ascent w.r.t.\ $\vtheta$ and variational parameters $\vxi$.}

    \label{fig:bound}
\end{figure}

\section{Background: Variational Inference for Gaussian Processes Models}
\label{sec:background}
Gaussian processes (GPs, \cite{rasmussen2006gaussian}) are distributions over functions, commonly used in machine learning to endow latent functions in generative models with rich and interpretable priors. These priors can provide strong inductive biases for regression tasks in the small data regime.
GP-based models are the ones that employ a GP prior over the (latent) functions $f(\cdot) \sim \GP(\mu(\cdot), \kappa(\cdot,\cdot'))$, where the prior is completely characterized by the mean function $\mu(\cdot)$ and covariance function $\kappa(\cdot, \cdot')$.
We denote $f(\cdot)$ as a function 
but occasionally simplify the notation to just $f$ in the interest of reducing clutter. 
Given a data set $\data = (\MX,\vy) =\{(\vx_i, y_i)\}_{i=1}^n$ of input--output pairs, we denote by $\vf$ the vector of function evaluations at the inputs  $\{f(\vx_i)\}_{i=1}^n$.
The function evaluation at $\vx_i$ are passed through likelihood functions to model the outputs $y \in \R$, \eg\ by specifying $p(\vy \mid \vf) \coloneqq \prod_{i=1}^{n} p(y_{i} \mid f_i)$. 
Prediction at a new test input $\vx_*$ is obtained by computing the distribution $p(f(\vx_*)\mid \data, \vx_*)$. For Gaussian likelihoods $\N(y_i \mid f_i, \sigma^2)$, the predictive distribution is available in closed form as a Gaussian distribution $\mathrm{N}(f(\vx_*) \mid m_{\textrm{GPR}}(\vx_*), v_{\textrm{GPR}}(\vx_*))$ with mean and variance defined as follows: 
\begin{equation}
   m_{\textrm{GPR}}(\vx_*) \coloneqq \vk_{\vf*}^\top ( \MKff + \sigma^2\MI_n )^{-1} \vy, \text{~~and~~} v_{\textrm{GPR}}(\vx_*) \coloneqq  \kappa_{**} - \vk_{\vf*}^\top ( \MKff + \sigma^2\MI_n )^{-1} \vk_{\vf*}, 
   \label{eq:gp_pred}
\end{equation}
where $\vk_{\vf*}$ is a vector of $\kappa(\vx_*, \vx_i)$ as the $i$\textsuperscript{th} element for all $\vx_i\in\MX$, $\MKff$ is an $n \times n$ matrix with $\kappa(\vx_i,\vx_j)$ as the $ij$\textsuperscript{th} entry, and $\kappa_{**} = \kappa(\vx_*, \vx_*)$.

\subsection{Variational Expectation--Maximization for GPs with Non-Conjugate Likelihoods}
\label{sec:em}
For non-Gaussian likelihoods, the posterior and predictive distributions are no longer Gaussian, and we need to resort to approximate inference methods. Variational inference is a popular choice because it allows for fast posterior approximation and hyperparameter learning via stochastic training~\citep{titsias2009variational, hensman2013gaussian}. Denoting kernel hyperparameters by $\vtheta$ and the corresponding GP prior by $p_{\vtheta}(\vf)$, the posterior distribution can be written as ${p_{\vtheta}(\vf \mid \vy) = p_{\vtheta}(\vf) \, p(\vy \mid \vf)/p_{\vtheta}(\vy)}$, where $p_{\vtheta}(\vy)=\mint p_{\vtheta}(\vf,\vy)  \dee \vf$ is the marginal likelihood of the observations.
We seek to approximate $p_{\vtheta}(\vf\mid\vy) \approx q_{\vf}(\vf)$ by a Gaussian distribution whose parameters can be obtained by optimizing the following evidence lower bound (ELBO) to the log-marginal likelihood,
\begin{equation}
\label{eq:elbo}
  \log p_{\vtheta}(\vy) \ge \cL_q(q_{\vf}, \vtheta) = \msum_{i=1}^n \Exp{q_{\vf}(f_i)}{\log p(y_i\mid f_i)} - \KL{q_\vf(\vf)}{p_{\vtheta}(\vf)}\,.
\end{equation}
For the variational approximation, it is a standard practice to choose the mean-covariance parameterization, denoted by $\vxi = (\vm, \MS)$.
It is also common to use the Cholesky factor $\ML$ instead \citep{challis2013gaussian} since it is uniquely determined for a covariance matrix.
The multivariate normal distribution is part of the exponential family \cite{wainwright2008}, i.e. it's probability density function take the form ${p(\vx) = \exp(\veta^\top \MT(\vx) - a(\veta))}$, with natural parameters $\veta = (\MS^{-1}\vm,\, -\MS^{-1}/2)$ and sufficient statistics $\MT(\vx)=[\vx, \vx\vx^\top]$. This natural parameterization is also a common choice, along with the associated expectation parameterization ${\vmu = \Exp{q}{\MT(\vx)}= (\vm,\, \MS + \vm\vm^\top)}$.
A final choice of the parameterization, called the \emph{whitened} parameterization~\cite{van2020framework}, uses a variable $\vv \sim \N(\vv; \vm_{\vv}, \MS_{\vv})$ along with the transformation $\vf = \ML \vv$, to parameterize $\vm =  \ML \vm_\vv$ and $\MS = \ML \MS_\vv \ML^\top$. 
 
In the following, we will consider the ELBO with several parameterizations and, to make the notation clearer, we will indicate the parameterization used with a subscript with $\cL$. For example, we may have  $\cL_{\xi}(\vxi, \vtheta) = \cL_{\eta}(\veta, \vtheta) = \cL_{\mu}(\vmu, \vtheta)$, which are all clearly equal due to a unique mapping between the parameterizations; see \citep{malago2015information} for more details on the maps.

The ELBO can be optimized by using a variational expectation--maximization (VEM) procedure, where we alternate between optimizing variational parameters, say $\vxi$, and hyperparameters $\vtheta$,
\begin{align}
   \text{E-step:}\quad \vxi^*_t &\leftarrow \textstyle\arg\max_{\vxi}\cL_{\xi}(\vxi, \vtheta_t),
   &&  \text{M-step:}\quad \vtheta_{t+1}\leftarrow \textstyle\arg\max_{\vtheta}\cL_{\xi}(\vxi_t^*, \vtheta), \label{eq:m-step}
\end{align}
where $t$ denotes the iterations, and we have explicitly written the dependence of optimal parameter $\vxi^*(\vtheta_t)$ as a function of the old parameter $\vtheta_t$. Both and E and M-steps can be carried out with gradient descent, for example, using an iteration of the form  ${\vxi_{t}^{(k+1)} \leftarrow \vxi_{t}^{(k)} + \rho_k \nabla_{\vxi} \cL_{\xi}(\vxi_{t}^{(k)}, \vtheta_t)}$ for E-step, which would ultimately converge to $\vxi_t^*$. A similar iterative method can be used for the M-step.

\subsection{Inference via Natural-Gradient Descent (NGD)}
\label{sec:natgrads}
A popular strategy for the E-step is to use natural gradient descent where we replace the gradient by the one preconditioned using the Fisher information matrix $\MF(\vxi)$ of $q_{\vf}(\vf)$. We denote natural gradients by $\tilde{\nabla}_{\vxi}\cL_{\xi}(\vxi, \vtheta) = \MF(\vxi)^{-1}\nabla_{\vxi} \cL_{\xi}(\vxi, \vtheta)$, to get the following update,
\begin{equation}
\label{eq:standard_natgrad}
    \vxi_{t}^{(k+1)} \leftarrow \vxi_{t}^{(k)} + \rho_k
    {\tilde{\nabla}_{\vxi}\cL_{\xi}(\vxi_{t}^{(k)}, \vtheta_t)} .
\end{equation}
Such updates can converge faster than gradient descent \citep{salimbeni2018natural, khan2017conjugate, khan2018}, and at times are less sensitive to the choice of the learning rate $\rho_k$ due to the scaling with the Fisher information matrix.
The implementation simplifies greatly when using natural parameterizations,
\begin{equation}
\label{eq:natural_natgrad}
\veta_{t}^{(k+1)} \leftarrow \veta_{t}^{(k)} + \rho_k \nabla_{\vmu}\cL_{\mu}(\vmu_{t}^{(k)}, \vtheta_t),
\end{equation}
because $\tilde{\nabla}_{\veta} \cL_{\eta}(\veta, \vtheta)= \nabla_{\vmu}\cL_{\mu}(\vmu, \vtheta)$, 
that is, the natural gradients with respect to $\veta$ are in fact the gradients with respect to the expectation parameter $\vmu$ \cite{khan2018}. 
In the remainder of the paper we will frequently use this property, and refer to the natural gradient with respect to $\veta$ by the gradients with respect to expectation parameterization $\vmu$. 
The VEM procedure with NGD has recently become a popular choice for sparse variants of GPs, which we explain next. 

\subsection{Sparse Variational GP Methods and Their Challenges}

Inference in GP models, whether conjugate or non-conjugate, suffers from an $\order(n^3)$ computational bottleneck required to invert the posterior covariance matrix.
A common approach to reduce the computational complexity is to use a sparse approximation relying on a small number $m \ll n$ representative inputs, also called \emph{inducing} inputs, denoted by $\MZ \coloneqq (\vz_1,\vz_2,\ldots,\vz_m)$~\cite{seeger2003bayesian, csato2002gaussian, quinonero2005unifying, williams2001using}.
Sparse variational GP methods~\cite{titsias2009variational, hensman2013gaussian,Hensman+Matthews+Ghahramani:2015, cheng_2016,Cheng+Boots:2017,salimbeni18:orthogonally} rely on a Gaussian approximation $q_{\vu}(\vu)$ over the functions $\vu = (f(\vz_1), f(\vz_2), \ldots, f(\vz_m))$ to approximate the posterior over arbitrary locations,
\begin{equation}
\label{eq:svgp_marginals}
  q_{\vu,\vtheta}(f(\cdot))= \mint p_{\vtheta}(f(\cdot) \mid \vu) \, q_{\vu}(\vu) \, \dee \vu , 
\end{equation}
where $p_{\vtheta}(f(\cdot) \mid \vu)$ is a conditional of the GP prior.
For example, for a Gaussian $q_{\vu}(\vu) = \N(\vu; \vm_{\vu}, \MS_{\vu})$, the posterior marginal of $f_i = f(\vx_i)$ takes the following form,
\begin{equation}
    q_{\vu, \vtheta}(f_i) 
    =  \N\left(f_i \mid \va^\top_i \vm_\vu, \kappa_{ii} - \va_i^\top (\MKuu-\MS_{\vu})\va_i  \right),
    \label{eq:SVGPmarginal}
\end{equation}
where ${\va_i = \MKuu\inv\vk_{\vu i}}$ with $\MKuu$ as the prior covariance evaluated at $\MZ$, and $\vk_{\vu i}$ as an $m$-length vector of $\kappa(\vz_j, \vx_i), \forall j$. The parameters $\vxi_\indu = (\vm_\indu, \MS_\indu)$ can be learned via an ELBO similar to \cref{eq:elbo}, 
\begin{equation}
   \cL_\xi(\vxi_\indu, \vtheta) \coloneqq \msum_{i=1}^n \Exp{q_{\vu, \vtheta}(f_i)}{ \log p(y_i \mid f_i) }  - \KL{q_\indu(\vu)}{p_{\theta}(\vu)}. 
   \label{eq:elbo_svgp_new}
\end{equation}
The variational objective for such sparse GP posteriors can be evaluated at a cost $\order(nm^2 + m^3)$, and optimization can be performed in $\order(m^3 + n_\mathrm{b} m^2)$ per iteration via stochastic natural-gradient methods with mini-batch size $n_\mathrm{b}$~\cite{hensman2013gaussian}.
This formulation also works for general likelihood functions~\cite{Hensman+Matthews+Ghahramani:2015}. This and the low computational complexity has lead to a wide adoption of the SVGP algorithm.
It is currently the state-of-the-art for sparse variants of GP and is available in the existing software implementations such as GPflow~\cite{GPflow:2017} and GPyTorch~\cite{GPyTorch}.

Despite their popularity, the current implementations are cumbersome and there is plenty of room for improvements.
For example, the methods discussed in \citet{salimbeni2018natural} (see \cref{app:pseudo-q-SVGP} for a summary) rely on the mean-covariance parameterization and NGD is performed in $\veta$-space. 
However, the natural gradients are implemented via chain rule: $(\nabla_{\vmu}\vxi) \nabla_{\vxi}\cL_\xi(\vxi, \vtheta)$ utilizing the gradients in the $\vxi$-space, which requires computation of additional Jacobians and multiplication operations.
Since other operations are done via mean-covariance parameterization, we need to go back and forth between $\veta$ and $\vxi$, which further increases the cost.
In addition, many existing implementations currently compute the natural gradient of the \emph{whole} ELBO, including the KL term which is not required; see \citet[Sec. 2.2]{khan2021BLR}.
Finally, the M-step is dependent on the choice of the parameterization used in E-step and can affect the convergence speed. To the best of our knowledge, this has not been investigated in the literature.

In what follows, we argue to use a \emph{dual} parameterization instead of the usual mean-covariance parameterization, and show that this not only simplifies computations of natural-gradients, but also gives rise to a tighter bound for hyperparameter learning and speed-up the whole VEM procedure.

\section{The $t$-VGP Method: Dual-Parameter Based Learning for GPs}
We start with a property of the optimal $q_{\vf}^*$ of \cref{eq:elbo}. \citet[Eq.~(18)]{khan2018} show, that $q_{\vf}^*$ can be parameterized by 2D vectors $\vlambda_i^* =(\lambda_{1,i}^*, \lambda_{2,i}^*)$ used in site functions $t_i^*(f_i)$,
\begin{equation}
    q^*_{\vf}(\vf) \propto p_{\vtheta}(\vf) \mprod_{i=1}^n \underbrace{ e^{\langle \vlambda_i^*, \MT(f_i)\rangle} }_{t_i^*(f_i)}, \textrm{ where } \vlambda_i^* = \nabla_{\vmu_i} \mathbb{E}_{q_{\vf}^*(f_i)}[ \log p(y_i \mid f_i) ].
   \label{eq:site_param}
\end{equation}
The vectors $\vlambda_i^*$ are equal to the natural gradient of the expected log-likelihood, where the expectation is taken with respect to the posterior marginal $q_{\vf}^*(f_i) = \N(f_i;m_i^*, S_{ii}^*)$ with $S_{ii}^*$ as the $i$'th diagonal element of $\MS^*$, and the gradient is taken with respect to its expectation parameter $\vmu_i = (m_i, m_i^2 + S_{ii})$ and evaluated at $\vmu_i^*$.
The vector $\MT(f_i) = (f_i, f_i^2)$ are the sufficient statistics of a Gaussian.
The $q_{\vf}^*$ uses local unnormalized Gaussian \emph{sites} $t_i^*(f_i)$, similarly to those used in the Expectation Propagation (EP) algorithm \cite{minka2001expectation}.
The difference here is that the site parameters $\vlambda_i^*$ are equal to natural gradients of the expected log-likelihood which are easy to compute using the gradient with respect to $\vmu_i$. 

The parameters $\vlambda_i^*$ can be seen as the optimal \emph{dual} parameters of a Lagrangian function with moment-matching constraints. We can show this in two steps:
\begin{enumerate}
    \item For each $p(y_i | f_i)$, we introduce a \emph{local} Gaussian $\widetilde{q}_i(f_i; \widetilde{\vmu}_i)$ with expectation parameters $\widetilde{\vmu}_i$.
    \item Then, we aim to match $\widetilde{\vmu}_i$ with the marginal moments $\vmu_i$ of the \emph{global} Gaussian $q_{\vf}(\vf; \vmu)$.
\end{enumerate}
This is written below as a Lagrangian where the middle term (shown in red) `decouples' the terms using the local Gaussians from those using the global Gaussian,
\begin{equation}
    \mathcal{L}_{\text{Lagrange}}(\vmu, \widetilde{\vmu}, \vlambda) = \msum_{i=1}^n \mathbb{E}_{\widetilde{q}_i(f_i;\widetilde{\vmu}_i)} [\log p(y_i \mid f_i)] 
    - {\color{red} \msum_{i=1}^n \langle \vlambda_i, \widetilde{\vmu}_i - \vmu_i \rangle }
    - \KL{q_\vf(\vf; \vmu)}{p_{\vtheta}(\vf)}.
    \label{eq:lagrangian}
\end{equation}
The parameter $\vlambda_i$ is the Lagrange multiplier of the moment-matching constraint $\widetilde{\vmu}_i = \vmu_i$ (here, $\vlambda$ and $\widetilde{\vmu}$ denote the sets containing all $\vlambda_i$ and $\widetilde{\vmu}_i$).
The optimal $\vlambda_i$ is equal to $\vlambda_i^*$ shown in \cref{eq:site_param}.
We can show this by, first setting the derivative with respect to $\vlambda_i$ to 0, finding that the constraints $\widetilde{\vmu}_i^* = \vmu_i^*$ are satisfied.
Using this and by setting the derivative with respect to $\widetilde{\vmu}_i$ to 0, we see that the optimal $\vlambda_i^*$ is in fact equal to the natural gradients, as depicted in \cref{eq:site_param}.

The optimal natural parameter of $q_{\vf}^*(\vf)$, denoted by $\veta^*$, has an `additive' structure that, as we will show, can be exploited to speed-up learning. The structure follows by setting derivatives w.r.t.\ $\vmu$ to 0,
\begin{equation}
    \nabla_{\vmu} \KL{q_\vf(\vf; \vmu^*)}{p_{\vtheta}(\vf)} = \vlambda^* 
    \quad \implies \quad 
    \veta^* = \veta_0(\vtheta) + \vlambda^*.
    \label{eq:lagrange_opt}
\end{equation}
The second equality is obtained by using the result that $\nabla_{\vmu} \KL{q(\vf;\vmu)}{p_{\vtheta}(\vf)} = \veta - \veta_0(\vtheta)$ where $\veta$ and $\veta_0(\vtheta)$ are natural parameters of $q(\vf;\vmu)$ and $p_{\vtheta}(\vf)$ respectively \cite[Sec.~2.2]{khan2021BLR}.
The final result follows by noting that the right-hand side is the natural parameter of $q_{\vf}^*(\vf)$ from \cref{eq:site_param}.
This implies that the global $q_{\vf}(\vf; \vmu^*) = q_{\vf}^*(\vf)$ is also equal to the optimal approximation, as desired. 

The Lagrangian formulation is closely related to the maximum-entropy principle \cite{jaynes1957} which forms the foundations of Bayesian inference \cite{jaynes1982}. Through moment matching, the prior is modified to obtain posterior approximations that explain the data well. Since $\vlambda_i^*$ are the optimal Lagrange multipliers, they measure the sensitivity of the optimal $q_{\vf}^*(\vf)$ to the perturbation in the constraints, and reveal the relative importance of data examples.
Therefore, the structure of the solution $\veta^*$ shown in \cref{eq:lagrange_opt} is useful for estimating $\vtheta$. The additive structure can be used to measure the relative importance of the prior to the dual parameters $\vlambda_i^*$. Our main idea is to use the structure to speed-up learning for SVGPs.

\cref{eq:lagrange_opt} can be rewritten in terms of the mean-covariance parameterization, to gain further insight about the structure. To do so, we use Bonnet and Price's theorem, and rearrange to get the following (see Eqs.~10 and 11 in \cite{khan2021BLR} for a similar derivation),
\begin{align}
  \vm^{*} &= - \MK_{\vf\vf} \valpha^{*},  &&\text{where $\valpha^*$ is a vector of }\: \alpha^{*}_{i} = \myexpect_{q_{\vf}^*(f_i)} \sqr{  \nabla_{f} \log p(y_i \mid f_i) },  \label{eq:var_m}\\
  (\MS^{*})^{-1} &= \MK_{\vf\vf}^{-1} + \diag(\vbeta^*), &&\text{where $\vbeta^*$ is a vector of }\:  \beta^*_i = \myexpect_{q_{\vf}^*(f_i)} \sqr{ -\nabla_{ff}^2 \log p(y_i \mid f_i) }.  \label{eq:var_V}
\end{align}
The variables $\alpha_i^*$ and $\beta_i^*$ can be easily obtained by using the gradient and Hessian of the log-likelihood, and using those we can get $\vlambda_i^* = (\beta_i^* m_i^* + \alpha_i^*, \,\, -\frac{1}{2}\beta_i^*)$.

Several other works have discussed such parameterizations, although our work is the first to connect it to natural gradients as the optimal Lagrange multiplier.
The representation theorem by \citet{kimeldorf1971some} is perhaps the most general result, but \citet{csato2002sparse} were the first to derive such parameterization for GPs; see Lemma~1 in their paper. Their result is for \emph{exact} posteriors which is intractable while ours is for Gaussian approximations and easy to compute. A minor difference there is that their parameterizations use the integrals of likelihoods (instead of log-likelihoods) with respect to the GP prior (instead of the posterior), but we can also express them as \cref{eq:site_param} where $q^*_{\vf}(\vf)$ is replaced by the true posterior $p_{\vtheta}(\vf\mid\vy)$. 

Parameterization of the variational posterior similar to ours are discussed in \cite{nickisch2008approximations, opper2009variational}, but the one by \citet{khan2013fast} is the most similar.
They establish the first connection to duality for cases where ELBO is convex with respect to the mean-covariance parameterization.
\citet{khan2014decoupled} extends this to non-convex ELBO using the Lagrangian function similar to ours, but written with the mean-covariance parameterization to get the solutions shown in \cref{eq:var_m,eq:var_V}.
As shown earlier, their $(\valpha,\vbeta)$ parameterization is just a reparameterization of our $\vlambda$ parameterization.
Here, we argue in favour of our formulation which enables the reformulation in terms of site functions in \cref{eq:site_param} and also allows us to exploit the `additive' structure in \cref{eq:lagrange_opt} to speed up hyperparameter learning.
The mean-covariance parameterization does not have these features.

\subsection{Improved Objective for Hyperparameter Learning}
\label{sec:improved_hyp}

We will now discuss a method to speed-up VEM by using the dual parameterization. The key idea is to exploit the form given in \cref{eq:lagrange_opt} to propose a better objective for the M-step. 

Standard VEM procedures, such as those shown in \cref{eq:m-step}, iterate pairs of E and M steps which we here describe in the context of the dual parameterization. In the E-step, starting from a hyperparameter $\vtheta_t$, the optimal variational distribution $q^*_{\vf}(\vf)$ maximizing the ELBO in \cref{eq:elbo} is computed. For the dual parameterization, we get the optimal variational parameters $\veta_t^* = \veta_0(\vtheta_t) + \vlambda_t^*$. Here, the subscripts $t$ in $\veta_t^*$ and $\vlambda_t^*$ indicate the dependence of the E-step iterations on $\vtheta_t$, while $\veta_0(\vtheta_t)$ indicates a direct dependence of the prior natural parameter over $\vtheta_t$. The standard M-step would then be to use $\veta_t^*$ in the ELBO in \cref{eq:m-step} as shown below, while we propose an alternate procedure where the prior $\veta_0(\vtheta)$ is left free (shown in red):
\begin{align}
    \text{Standard M-step: }\quad \vtheta_{t+1} &= \argmin_{\vtheta} \cL_\eta(\veta_0(\vtheta_t) + \vlambda_t^*, \vtheta) \label{eq:standardM}\\  
    \text{Proposed M-step: }\quad \vtheta_{t+1} &= \argmin_{\vtheta} \cL_\eta(\veta_0({\color{red} \vtheta}) + \vlambda^*_t, \vtheta)\label{eq:proposedM}
\end{align}
This proposed objective is still a lower bound to the marginal likelihood and it corresponds to the ELBO in \cref{eq:elbo} with a distribution whose natural parameter is
$\hat{\veta}_t(\vtheta) = \veta_0({\color{black} \vtheta}) + \vlambda^*_t$
and thus depends on $\vtheta$. We denote this distribution by $q_{\vf}(\vf; \hat{\veta}_t(\vtheta))$.
The ELBO is different from the one the distribution obtained after the E-step, with natural parameter ${\veta_t^*}$ which is independent of $\vtheta$. We denote this distribution by $q_{\vf}(\vf; \veta_t^*)$.
Clearly, at $\vtheta=\vtheta_t$ both objectives match, and so do their gradient with respect to $\vtheta$, but they generally differ otherwise. We argue that the proposed M-step could lead to a tighter lower bound; see \cref{fig:bound} for an illustration.

In the standard M-step, the dependency of the bound on $\vtheta$ is only via the KL divergence in \cref{eq:elbo}. In the M-step we propose, this dependency is more intricate because the expected log-likelihood also depend on $\vtheta$. Yet, as we show now, it remains simple to implement. The lower bound in the proposed M-step takes a form where an existing implementation of GP regression case can be reused.
\begin{equation}
    \cL_\eta(\veta_0(\vtheta) + \vlambda_t^*, \vtheta) 
    = \myexpect_{q_{\vf}(\vf;\hat{\veta}_t(\vtheta))} \sqr{ \log \frac{\prod_{i=1}^n p(y_i \mid f_i) \cancel{ p_{\vtheta}(\vf) } }{ \frac{1}{\mathcal{Z}_t(\vtheta)} \prod_{i=1}^n t_i^*(f_i) \cancel{p_{\vtheta}(\vf)} } }
    = \log \mathcal{Z}_t(\vtheta) + {c(\vtheta)} ,
\end{equation}
where 
$c(\vtheta)=\msum_{i=1}^n \myexpect_{q_{\vf}(\vf;\hat{\veta}_t(\vtheta))} \sqr{ \log \frac{p(y_i \mid f_i)}{t_i^*(f_i)} }$ and
$\log \mathcal{Z}_t(\vtheta)$ is the log-partition of $q_{\vf}(\vf;\hat{\veta}_t(\vtheta))$, 
\begin{equation}
        \log \mathcal{Z}_t(\vtheta) = -\tfrac{n}{2} \log (2\pi) - \tfrac{1}{2} \log| \diag(\vbeta^*_t)^{-1} + \MK_{\vf\vf}(\vtheta)| - \tfrac{1}{2} \widetilde{\vy}^\top \sqr{ \diag(\vbeta^*_t)^{-1} + \MK_{\vf\vf}(\vtheta) }^{-1} \widetilde{\vy}.
        \label{eq:Zt_VGP}
\end{equation}
Here, $\widetilde{\vy}$ is a vector of $\widetilde{y}_i = -\frac{1}{2} \lambda_{1,i}^*/\lambda_{2,i}^*$, and we have explicitly written $\MK_{\vf\vf}(\vtheta)$ to show its direct dependence on the hyperparameter $\vtheta$.
The gradients of $\mathcal{Z}_t(\vtheta)$ can be obtained using GP regresssion code, while the gradient of $c(\vtheta)$ can be obtained using standard Monte-Carlo methods.
A similar lower bound was originally used in the implementation\footnote{See \url{https://github.com/emtiyaz/cvi/blob/master/gp/infKL_cvi.m}} provided by \citet{khan2017conjugate}, but they did not use it for hyperparameter learning.

For GP regression, we recover the exact log-marginal likelihood $\log p_{\vtheta}(\vy \mid \data)$ for all values of $\vtheta$. Indeed $\vlambda_i^* = (y_i/\sigma^2, -1/(2\sigma^2) )$, which means that the sites exactly match the likelihood terms so $c(\vtheta)=0$. This also gives us $\widetilde{y}_i = y_i$ and $\beta_i = 1/\sigma^2$, and we get $\log \mathcal{Z}_t(\vtheta) = \log p_{\vtheta}(\vy \mid \data)$. 

For non-conjugate problems, we found it to be tighter bound than the standard ELBO (\cref{eq:standardM})) which could speed-up the procedure.
This is illustrated in \cref{fig:fixed_point_classif} (top row) where the proposed ELBO is compared to two other parameterizations (mean-covariance and whitened) for many values of $\vtheta_t = \vtheta_{\text{old}}$. 
We see that the maximum value (shown with a dot) remains rather stable for the proposed method compared to the other two. This is as expected due to \cref{eq:proposedM} where we expect the solutions to become less sensitive to $\vtheta_t$ because we have replaced $\veta_0(\vtheta_t)$ by $\veta_0(\vtheta)$.
The bottom row in \cref{fig:fixed_point_classif} shows the iterative steps $(\vtheta_{t+1}, \vtheta_t)$ for a few iterations, where we see that, due to the stable solutions of the new ELBO, the iterations quickly converge to the optimum. 
Exact theoretical reasons behind the speed-ups are currently unknown to us. 
We believe that the conditioning of the ELBO is improved under the new parameterization. We provide some conditions in \cref{app:tighter_bound} under which the new ELBO would provably be tighter.

\subsection{Faster Natural Gradients for Inference Using the Dual Paramterization}
So far, we have assumed that the both E and M steps are run until convergence, but it is more practical to use a stochastic procedure with partial E and M steps, for example, such as those used in \cite{hensman2013gaussian, hoffman13a}.
Fortunately, with natural-gradient descent, we can ensure that the iterations also follow the same structure as that of the solution shown in \cref{eq:site_param} and \cref{eq:lagrange_opt}. 
Specifically, we use the method of \citet{khan2017conjugate}, expressed in terms of the dual parameters $\vlambda$ and natural gradients of the expected log-likelihoods 
$\vg_i^{(k)} = {\nabla}_{\vmu_i} \mathbb{E}_{q_{\vf}^{(k)}(f_i)}[ \log p(y_i \mid f_i)]$ 
(see also \cite[Sec.~5.4]{khan2021BLR}),
\begin{equation}
    q^{(k+1)}_{\vf}(\vf) \propto p_{\vtheta}(\vf) \mprod_{i=1}^n \underbrace{ e^{\langle \vlambda_i^{(k+1)}, \MT(f_i)\rangle} }_{t_i^{(k+1)}(f_i)}, \textrm{ where } \vlambda_i^{(k+1)} = (1-r_k)\vlambda_i^{(k)} + r_k \vg_i^{(k)} .
    \label{eq:ngd_khanandlin}
\end{equation}
The convergence of these iterations is guaranteed under mild
conditions discussed in \cite{khan2016faster}.
The natural parameter of $q_{\vf}^{(k)}(\vf)$ at iteration $k$ can be written in terms of $\vtheta$ as follows,
\begin{equation}
\veta^{(k)} = \veta_0(\vtheta) + \vlambda^{(k)},
\end{equation}
and the expectation parameters $\vmu_i^{(k)}$, required to compute the natural gradients of the expected log-likelihood, can be obtained by using a map from the natural parameter $\veta^{(k)}$. 

The updates hold for any $\vtheta$ and can be conveniently used as $\vtheta = \vtheta_t$ at the E-step of the $t$\textsuperscript{th} EM iteration. 
We name $t$-VGP the EM-like algorithm with 1) an E-step consisting of the natural gradient updates of \cref{eq:ngd_khanandlin}, and, 2) the proposed M-step introduced in \cref{eq:proposedM}.

\section{The $t$-SVGP Method: Dual-Parameter Based Inference for SVGP}
\label{sec:svgp}

We now extend the dual parameterization based stochastic VEM procedure to the SVGP case and refer to the resulting algorithm as $t$-SVGP. The optimality property shown in \cref{eq:site_param} is shared by the ELBO given in \cref{eq:elbo_svgp_new}. That is, we can express the optimal $q_{\vu}^*(\vu)$ in terms of $n$ 2D parameters $\vlambda_i^*$, 
\begin{equation}
    q^*_{\vu}(\vu) \propto p_{\vtheta}(\vu) \mprod_{i=1}^n \underbrace{ e^{\langle \vlambda_i^*, \MT(\va_i^\top \vu) \rangle} }_{t_i^*(\vu)}, \textrm{ where } \vlambda_i^* = \nabla_{\vmu_{\indu, i}} \mathbb{E}_{q_{\vu}^*(f_i)}[ \log p(y_i \mid f_i) ].
   \label{eq:site_param_svgp}
\end{equation}
The difference here is that the site parameters use the sufficient statistics $\MT(\va_i^\top \vu)$, defined via the projections $\va_i = \MKuu\inv\vk_{\vu i}$. 
The natural gradients of the expected log-likelihood are computed by using the marginal $q_{\vu}^*(f_i)$ defined in \cref{eq:svgp_marginals} by using $\vxi_{\vu}^* = (\vmu_\indu^*, \MS_\indu^*)$ evaluated at the expectation parameters $\vmu_{\indu, i}^*$. Note that both $\va_i$ and $q_{\indu}^*(f_i)$ depend on $\vtheta$, but we have suppressed the subscript for notation simplicity. 

Similarly to the VGP case, the $\vlambda_i^*$ are the optimal dual parameters that measure the sensitivity of the solution to the perturbation in the moments of the posterior marginal $q_{\vu}^*(f_i)$. This suggests that we can design a similar VEM procedure that exploits the structure of solution in \cref{eq:site_param_svgp}. The structure is shown below in terms of the natural parameterization of $q^*_{\vu}(\vu)$ for sufficient statistics $\MT(\vu)$,
\begin{equation}
    \rnd{\MS_{\vu}^*}\inv\vm_{\vu}^* = \MKuu^{-1} \underbrace{ \rnd{ \msum_{i=1}^{n} \vk_{\vu i} {\lambda}_{1,i}^*} }_{= \bar{\vlambda}_1^*}
    \text{ and }
    \rnd{\MS_{\vu}^*}\inv = \MKuu\inv + \MKuu\inv \underbrace{ \rnd{ \msum_{i=1}^{n} \vk_{\vu i} {\lambda}_{2,i}^* \vk_{\vu, i}^\top }}_{= \bar{\MLambda}_2^*} \MKuu\inv,
    \label{eq:lambda_mxm}
\end{equation}
the quantities $\MKuu$ and $\vk_{\vu i}$ directly depend on $\vtheta$ and we can express the ELBO as the partition function of a Gaussian distribution, similarly to \cref{eq:Zt_VGP} (exact expression in \cref{app:proposed_elbo}). 
For large data sets, storing all the $\{\vlambda_i^*\}_{i=1}^n$ might be problematic, and we can instead store only $\bar{\vlambda}^*_1$, a $m$-length vector, and $\bar{\MLambda}^*_2$, a $m\times m$ matrix.
This \emph{tied} parameterization is motivated from the \emph{site-tying} setting in sparse EP \cite{bui2017unifying, li2015stochastic} where the goal is to reduce the storage. 
The parameterization ignores the dependency of $\vk_{\vu i}$ over $\vtheta$ and may reduce the coupling between $\vtheta$ and $q_{\vu}^*$, but it is suitable for large data sets. An alternative tying method consists in storing the sums and the flanking $\MKuu\inv$ terms.

We detail the final stochastic variational procedure which we refer to as $t$-SVGP: 
Given a parameter $\vtheta_t$, we run a few iterations of the E-step. 
At each iteration $k$ of the E-step, given $\vm_\indu^{(k)}$ and $\MS_\indu^{(k)}$, we sample a minibatch $\mathcal{M}$ and compute the natural gradients by first computing
\[\alpha_i^{(k)} = \myexpect_{q_{\vu}^{(k)}(f_i)} \sqr{  \nabla_{f} \log p(y_i \mid f_i) } \quad \textrm{ and }  \quad \beta_i^{(k)} = \myexpect_{q_{\vu}^{(k)}(f_i)} \sqr{ -\nabla_{ff}^2 \log p(y_i \mid f_i) } , \]
using the marginals $q_{\vu}^{(k)}(f_i)$ from \cref{eq:svgp_marginals}.
The natural gradients of the expected log-likelihood for the $i$\textsuperscript{th} site is then equal to ${\vg_i^{(k)} = (\beta_i^{(k)} m_i^{(k)} + \alpha_i^{(k)}, \,\, \beta_i^{(k)})}$.
Using these natural gradients we can use an iterative procedure similar to \cref{eq:ngd_khanandlin} but now on the tied parameters, 
\begin{align}
    \bar{\vlambda}_1^{(k+1)} &\leftarrow (1-r_k) \bar{\vlambda}_1^{(k)} + r_k \msum_{i \in \mathcal{M}} \vk_{\vu i} \vg_{1,i}^{(k)},\\
    \bar{\MLambda}_2^{(k+1)} &\leftarrow (1-r_k) \bar{\MLambda}_2^{(k)} + \, r_k \msum_{i \in \mathcal{M}} \vk_{\vu i}\vk_{\vu i}^\top \vg_{2,i}^{(k)} .
\end{align}
The natural parameter required can be obtained using \cref{eq:lambda_mxm},  
\begin{equation}
    \MS_{\vu}^{(k)} \leftarrow \rnd{ \MKuu\inv + \MKuu\inv \bar{\MLambda}_2^{(k)} \MKuu\inv }^{-1} \quad \text{and} \quad 
    \vm_{\vu}^{(k)} \leftarrow \MS_{\vu}^{(k)} \MKuu^{-1} \bar{\vlambda}_1^{(k)}.
\end{equation}

After a few E-steps, we update the parameters with a gradient descent step using the gradient of the log-partition function of $q_{\vf}^{(k)}(\vu)$ with respect to $\vtheta$.
In \cref{sec:computation}, we detail how to efficiently make predictions and compute the ELBO under parameterization \cref{eq:lambda_mxm}. The full algorithm is given in \cref{app:pseudo-t-SVGP}.
The convergence of the sequence of stochastic updates in the E-step is guaranteed under mild conditions discussed in \cite{khan2016faster} for the untied setting. Site-tying introduces a bias but does not seem to affect convergence in practice.

\begin{figure}[t]
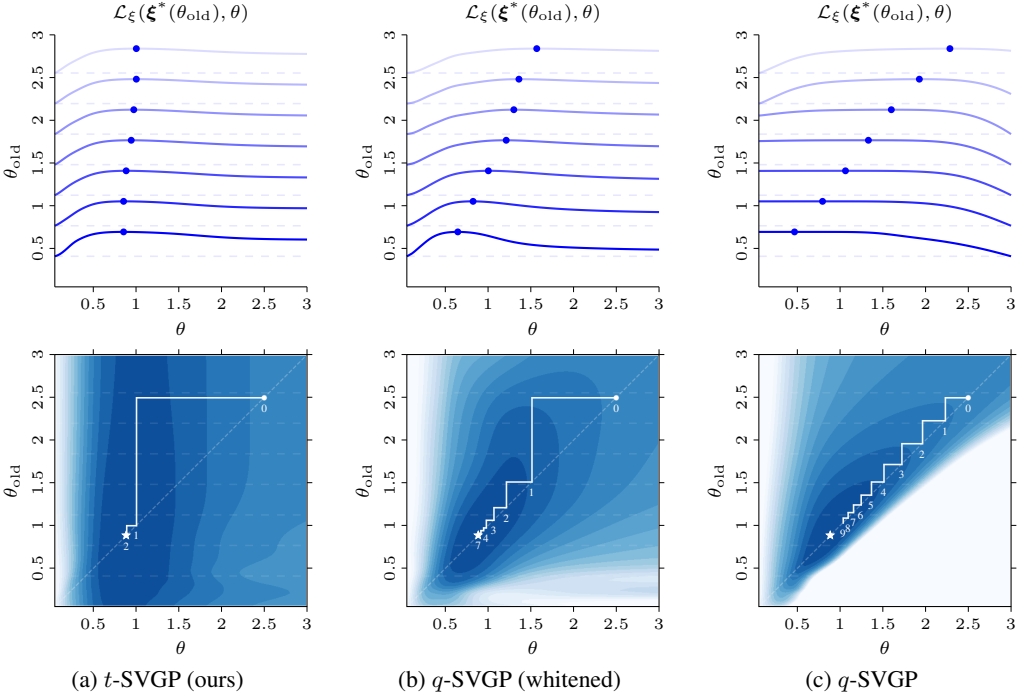

  \centering\scriptsize
  \setlength{\figurewidth}{.24\textwidth}
  \setlength{\figureheight}{\figurewidth}
  \pgfplotsset{axis on top,scale only axis,y tick label style={rotate=90},xtick={0.5,1,1.5,2,2.5,3},ytick={0.5,1,1.5,2,2.5,3}, x tick label style={font=\tiny},y tick label style={font=\tiny},title style={yshift=-4pt},major tick length=2pt} 
  \pgfplotsset{axis y line*=left, axis x line*=bottom}
  \begin{subfigure}[t]{.33\textwidth}
    \centering

\\[-3pt]
    \caption{$t$-SVGP (ours)}
  \end{subfigure}
  \hfill
  \begin{subfigure}[t]{.33\textwidth}
    \centering
%
\\[-3pt]
    \caption{$q$-SVGP (whitened)}    
  \end{subfigure}
  \hfill
  \begin{subfigure}[t]{.33\textwidth}
    \centering
%
\\[-3pt]
    \caption{$q$-SVGP}
  \end{subfigure}  
  \caption{
EM iterations for hyperparameter learning on a classification task on a a toy dataset.
{\bf Top row:} for the M-step, the optima of the $t$-SVGP objective (left) are less sensitive to the initial hyperparameter value $\vtheta_\mathrm{old}$, compared to $q$-SVGP (middle and right). 
{\bf Bottom row:} the EM iterations are shown in white on top of the EM objectives for a range of starting values for $\vtheta_\mathrm{old}$. 
The starting point $\theta = 2.5$ is marked with a white dot and the optimum with a star. %
$t$-SVGP converges much faster (2 iterations in leftmost plot) compared to $q$-SVGP (middle and rightmost plots which take ${>}5$ iterations). %
  }
  \label{fig:fixed_point_classif}
\end{figure}

\section{Empirical Evaluation}
\label{sec:experiments}
We conduct experiments to highlight the advantages of using the dual parameterization. Firstly, we study the effects of the improved objective for hyperparameter learning of $t$-SVGP versus $q$-SVGP. We study the objective being optimized for a single M-step, after an E-step ran until convergence. We then show a full sequence of EM iterations on small data sets. For large-scale data, where running steps to convergence is expensive, we use partial E and M-steps and mini-batching.
Our improved bound and faster natural gradient computations show benefits in both settings. It is worth noting that the E-step for both $q$-SVGP with natural gradients and $t$-SVGP are identical up to machine precision, and any differences in performance are to be attributed to the different parameterization. 

\paragraph{The Role of the Learning Objective}
In \cref{fig:bound}, we learn the kernel hyperparameters $\vtheta$ in a GP classification task via coordinate ascent of the lower bound $\cL_{\xi}(\vxi, \vtheta)$, where $\vxi$ are the variational parameters, \ie\ via EM. Starting at hyperparameter $\vtheta_{\text{old}}$, we denote by $\vxi^*(\vtheta_{\text{old}})$ the associated optimal variational parameters. Updating $\vtheta$ consists in optimizing $\cL_{\xi}(\vxi^*(\vtheta_{\text{old}}), \vtheta)$ which we show on the left panel for the dual parameterization (blue), the standard whitened (orange) and unwhitened (green) SVGP parameterizations. The dual parameterization leads to a tighter bound and thus to bigger steps and faster overall convergence as shown on the right for the 
illustrative toy classification
task, starting at $(\theta_1, \theta_2) = (1,1)$, in the extreme case of taking both the E and M step to convergence. For the 
toy 
data set we use $m=10$ inducing points (see details in \cref{sec:datasets}).

In \cref{fig:fixed_point_classif}, we also use the 
toy 
data and parameters as in \cref{fig:bound}, but we show how the learning objective changes over iterations. 
The blue contours show, for all initial $\vtheta_{\text{old}}$, the objective maximized in the M-step, \ie\ $\cL_{\xi}(\vxi^*(\vtheta_{\text{old}}),\vtheta)$.
The orange lines show, for all initial $\vtheta_{\text{old}}$, the outcome of an E-step followed by and M-step, \ie\ $\vtheta^*(\vtheta_{\text{old}})= \argmax_{\vtheta}\cL_{\xi}(\vxi^*(\vtheta_{\text{old}}),\vtheta)$. 
The EM iterations converge to the fixed points of $\vtheta^*$, \ie\ its intersection with the diagonal line of the identity function.
A flatter line around the optimal value is more desirable as it means the iterations converge faster to the optimum value which in this experiment is just below one, while a line close to the diagonal leads to slow convergence.
Here $t$-SVGP has the fastest convergence, $q$-SVGP performs poorly, although whitening clearly helps the optimisation problem.
The dark dashed lines show how optimising $\theta$ would look starting from $\theta_0 = 2.5$ and running $8$ iterations for the different models.

\begin{figure}[t]
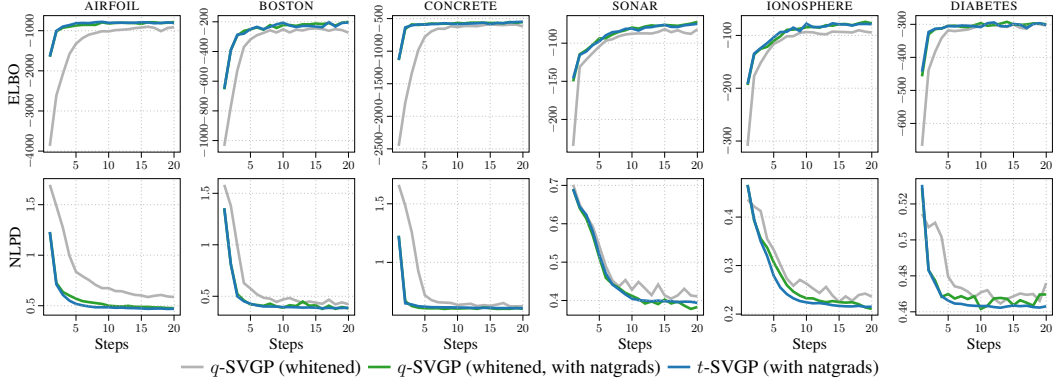

  \centering\scriptsize
  \setlength{\figurewidth}{.26\textwidth}
  \setlength{\figureheight}{\figurewidth}
  \pgfplotsset{axis on top,scale only axis,y tick label style={rotate=90}, x tick label style={font=\footnotesize},y tick label style={font=\footnotesize},title style={yshift=-4pt,font=\large}, y label style={font=\large},x label style={font=\large},grid=major}
  \begin{minipage}[t]{.17\textwidth}
    \raggedleft

\\[-2.5em]
  \caption{Comparison of convergence in terms of ELBO and negative log-predictive density (NLPD) as averages over 5-fold cross-validation runs, on UCI regression and classification tasks. All methods were trained (incl.\ hyperparameters and 50 inducing input locations) with matching learning rate. Natural gradient--based training is superior, and the $t$-SVGP parameterization improves stability.}
  \label{fig:uci}
\end{figure}

\paragraph{Evaluation on UCI Classification and Regression Tasks}
We use common small and mid-sized UCI data sets to test the performance of our method against $q$-SVGP with natural gradient optimisation and normal $q$-SVGP trained with Adam optimizer for the variational parameters. All methods use Adam for the hyperparameters. The exact details of the data sets can be found in \cref{sec:datasets}. Here we again take the approach that the optimal way to optimize the ELBO if computational budget allows is to alternate between performing E and M-steps till convergence. We plot how the different inference schemes perform for ELBO and NLPD on a hold test set. We perform 5-fold cross validation with the results in \cref{fig:uci} showing the mean of the folds for ELBO and NLPD. Natural gradient variants of $q$-SVGP clearly perform better than non natural gradient $q$-SVGP. Our method $t$-SVGP seems more stable specifically in NLPD for most data sets if not equal to $q$-SVGP. For $q$-SVGP, we have used the whitened version which as we noted helps with hyperparameter optimisation.

\paragraph{Improved Efficiency in Large-scale Inference}
To highlight {\em practical} benefits, we show the performance of our stochastic and sparse {$t$-SVGP} framework on the MNIST (\cite{lecun1998mnist}, available under CC BY-SA 3.0) multiclass-classification task (for details see \cref{sec:datasets}). Given the data set is $n = 70{,}000$, minibatching is needed and we adopt the parameterization of \cref{eq:lambda_mxm} meaning we match $q$-SVGP for parameter storage complexity. We compare against the natural gradient $q$-SVGP implementation but not the non-natural gradient version since it produces considerably worse performance. In large scale minibatching experiments performing full E and M steps may not be efficient. Instead we perform partial steps for both. A single E and M step can be thought of as the approach outlined in \cite{salimbeni2018natural}. All experiments are performed with a batch size of $n_\mathrm{b} = 200$ and $m=100$ inducing points and the optimization is ran until convergence using the Adam optimizer for the hyperparameters (M-step). \cref{tbl:comparison} shows different variations of learning rates and iterations of E- and M-steps. The results suggest some benefits in running partial EM steps. The $t$-SVGP formulation performs equally if not better than $q$-SVGP under all settings.

\setlength{\columnsep}{1em}
\setlength{\intextsep}{0pt}
\begin{wraptable}{r}{.48\textwidth}
  \raggedleft
  \scriptsize
  \caption{NLPD on MNIST benchmarks for different learning rates and E and M steps.}
  \label{tbl:comparison}  
  \setlength{\tabcolsep}{2pt}
  \renewcommand{\arraystretch}{.75}  
  \newcommand{\PreserveBackslash}[1]{\let\temp=\\#1\let\\=\temp}
  \newcolumntype{C}[1]{>{\PreserveBackslash\centering}m{#1}}
  \begin{tabularx} 
  {.48\textwidth}
  {C{.12\textwidth} C{.12\textwidth} | C{.05\textwidth} C{.05\textwidth} | C{.03\textwidth} C{.03\textwidth} }
     \toprule
     \multicolumn{2}{c}{\sc NLPD  } & \multicolumn{2}{c}{\sc LR} & \multicolumn{2}{c}{\sc Steps} \\
     \sc $q$-SVGP & \sc $t$-SVGP  & \sc E & \sc M & \sc \#E & \sc \#M \\
     \midrule
       $\bf0.304 {\pm} 0.015$ & $\bf0.304 {\pm} 0.006$ & $0.040$ & $0.05$ & $1$ & $1$  \\     
          $0.289 {\pm} 0.010$ & $\bf0.283 {\pm} 0.007$ & $0.035$ & $0.10$ & $2$ & $1$  \\
          $0.293 {\pm} 0.020$ & $\bf0.281 {\pm} 0.010$ & $0.030$ & $0.10$ & $3$ & $1$  \\
          $0.259 {\pm} 0.010$ & $\bf0.255 {\pm} 0.006$ & $0.025$ & $0.03$ & $4$ & $2$  \\
       $\bf0.282 {\pm} 0.007$ & $\bf0.283 {\pm} 0.006$ & $0.050$ & $0.03$ & $4$ & $2$  \\
          $0.243 {\pm} 0.003$ & $\bf0.230 {\pm} 0.009$ & $0.030$ & $0.03$ & $4$ & $1$  \\

     \bottomrule
  \end{tabularx}
\end{wraptable}
In \cref{fig:mnist}, we show the speed advantage of $t$-SVGP over $q$-SVGP due to cheaper natural gradient updates. We compare against the state-of-the-art implementation of SVGP in GPflow (\cite{GPflow:2017}, v2.2.1) and a closely matched implementation of our method in GPflow. We compare wall-clock time to compute 150 steps of the algorithm for both methods in terms of NLPD and ELBO taking single E and M-steps (MacBook pro, 2~GHz CPU, 16~GB RAM). 
Our implementation avoids the use of sluggish automatic differentiation to compute the natural gradients, and, even if our implementation is not as optimized as SVGP in GPflow, it is roughly 5 times faster on this standard benchmark.

\begin{figure}[t]
  \centering\scriptsize
  \setlength{\figurewidth}{.4\textwidth}
  \setlength{\figureheight}{.66\figurewidth}
  \pgfplotsset{axis on top,scale only axis,y tick label style={rotate=90},grid=major, legend style={fill=white}, every y tick scale label/.style={at={(rel axis cs:-0.1,.85)},anchor=south west,inner sep=1pt}}
  \begin{subfigure}[t]{.48\textwidth}
    \centering
\begin{tikzpicture}

\definecolor{color0}{rgb}{0.12156862745098,0.466666666666667,0.705882352941177}
\definecolor{color1}{rgb}{1,0.498039215686275,0.0549019607843137}

\begin{axis}[
height=\figureheight,
legend cell align={left},
legend style={
  fill opacity=1.0,
  draw opacity=1,
  text opacity=1,
  at={(0.97,0.03)},
  anchor=south east,
  draw=white!80!black
},
tick align=outside,
tick pos=left,
width=\figurewidth,
x grid style={white!69.0196078431373!black},
xlabel={Wall-clock time (s)},
xmin=-4.37717480368614, xmax=93.9438896700064,
xtick style={color=black},
y grid style={white!69.0196078431373!black},
ylabel={ELBO},
ymin=-159537.59543692, ymax=-15515.6214811108,
ytick style={color=black}
]
\addplot [thick, color0]
table {%
0.0919644905726115 -152160.399348714
0.183928981145223 -151159.81004415
0.275893471717834 -150126.605931226
0.367857962290446 -149531.11651346
0.459822452863057 -148990.869497116
0.551786943435669 -146397.392548599
0.64375143400828 -145534.552792069
0.735715924580892 -145481.361659615
0.827680415153503 -142503.747226941
0.919644905726115 -142395.299610957
1.01160939629873 -139274.087033763
1.10357388687134 -139445.37874139
1.19553837744395 -138267.652903007
1.28750286801656 -134239.377425973
1.37946735858917 -132940.672918364
1.47143184916178 -133462.934013889
1.5633963397344 -131944.87672564
1.65536083030701 -134415.588273761
1.74732532087962 -130640.283454118
1.83928981145223 -128507.697044811
1.93125430202484 -125328.097272419
2.02321879259745 -125399.117358124
2.11518328317006 -123192.214315543
2.20714777374268 -121749.739239938
2.29911226431529 -117183.568396951
2.3910767548879 -118671.426187801
2.48304124546051 -116829.008986772
2.57500573603312 -113133.791035223
2.66697022660573 -108595.763497298
2.75893471717835 -109638.814616812
2.85089920775096 -107797.71115507
2.94286369832357 -100252.720137008
3.03482818889618 -99723.6861636128
3.12679267946879 -98594.8392347584
3.2187571700414 -95961.2191840416
3.31072166061401 -97831.8787601753
3.40268615118663 -89257.6440269661
3.49465064175924 -89487.0579699556
3.58661513233185 -83567.4000716221
3.67857962290446 -81427.9872458299
3.77054411347707 -74497.2130312802
3.86250860404968 -76555.0530612692
3.95447309462229 -69177.6875299313
4.04643758519491 -70410.5049255651
4.13840207576752 -63032.9619537954
4.23036656634013 -60932.108416073
4.32233105691274 -55105.5311129129
4.41429554748535 -57558.1179543502
4.50626003805796 -52129.8403221226
4.59822452863057 -51639.165564187
4.69018901920319 -48274.8530013542
4.7821535097758 -49320.7059628979
4.87411800034841 -46891.0141809392
4.96608249092102 -44060.7304451426
5.05804698149363 -45210.4453932349
5.15001147206624 -44303.9002072651
5.24197596263885 -43250.4016888153
5.33394045321147 -39267.8390798195
5.42590494378408 -39819.0507703656
5.51786943435669 -38493.8332081088
5.6098339249293 -38470.0866741355
5.70179841550191 -37257.0709298723
5.79376290607452 -35398.7924766159
5.88572739664714 -37658.3762112412
5.97769188721975 -34329.6096072786
6.06965637779236 -35652.2377099493
6.16162086836497 -33269.9481238173
6.25358535893758 -33659.6708135508
6.34554984951019 -32101.2698567581
6.4375143400828 -31225.470499397
6.52947883065542 -31881.5924625136
6.62144332122803 -31383.8491232021
6.71340781180064 -33277.2861881268
6.80537230237325 -31985.3616053008
6.89733679294586 -31049.2657135713
6.98930128351847 -33168.4787497422
7.08126577409108 -31914.8036856128
7.1732302646637 -30453.7177375318
7.26519475523631 -32781.9448615402
7.35715924580892 -30179.4844905658
7.44912373638153 -31684.2811051938
7.54108822695414 -31040.6839998835
7.63305271752675 -29560.2135250225
7.72501720809937 -30966.3292280302
7.81698169867198 -31071.2454942471
7.90894618924459 -27075.1170090885
8.0009106798172 -28502.9657912153
8.09287517038981 -28215.4912756828
8.18483966096242 -31522.9226222572
8.27680415153503 -31555.8241580158
8.36876864210765 -30073.0384421175
8.46073313268026 -26435.6598732292
8.55269762325287 -28913.9549665487
8.64466211382548 -25145.0350855943
8.73662660439809 -26976.9528912097
8.8285910949707 -28265.8506921727
8.92055558554331 -28838.8353561145
9.01252007611593 -28877.4520389306
9.10448456668854 -27268.2424151118
9.19644905726115 -27790.9954471495
9.28841354783376 -29194.6512558641
9.38037803840637 -25588.3215079409
9.47234252897898 -25135.2286657261
9.5643070195516 -28026.9503981789
9.65627151012421 -27678.574550882
9.74823600069682 -28430.649652808
9.84020049126943 -30502.2210011378
9.93216498184204 -25155.6125132859
10.0241294724147 -28871.4619776544
10.1160939629873 -29535.3219830111
10.2080584535599 -28644.773428835
10.3000229441325 -24966.5243183568
10.3919874347051 -27470.8921754649
10.4839519252777 -27132.7343076959
10.5759164158503 -27414.6583704896
10.6678809064229 -26963.1840067992
10.7598453969955 -26587.4121960224
10.8518098875682 -24452.1711578485
10.9437743781408 -25332.7094583053
11.0357388687134 -25197.2989406628
11.127703359286 -27065.9113029224
11.2196678498586 -22062.0748427385
11.3116323404312 -26667.915615188
11.4035968310038 -27059.6766685038
11.4955613215764 -27885.4346392474
11.587525812149 -25166.7441239283
11.6794903027217 -28359.3678871101
11.7714547932943 -26407.7874993173
11.8634192838669 -26192.5423393454
11.9553837744395 -25412.5809617987
12.0473482650121 -26184.053402068
12.1393127555847 -26731.4015607398
12.2312772461573 -23212.8047411742
12.3232417367299 -26493.7925588467
12.4152062273026 -27128.2008725936
12.5071707178752 -27014.2871496033
12.5991352084478 -24238.778259574
12.6910996990204 -24023.9020537127
12.783064189593 -26457.0067272763
12.8750286801656 -23389.3670062211
12.9669931707382 -26820.1787419397
13.0589576613108 -24433.0242547909
13.1509221518834 -23187.3772947344
13.2428866424561 -25604.3064180417
13.3348511330287 -24764.3315900673
13.4268156236013 -23462.1935927393
13.5187801141739 -24072.8546409088
13.6107446047465 -24788.9954437151
13.7027090953191 -25436.8289349174
13.7946735858917 -25669.799007386
};
\addlegendentry{$t$-SVGP}
\addplot [thick, color1]
table {%
0.596498335838318 -152991.142075292
1.19299667167664 -151826.749539742
1.78949500751495 -150426.614089789
2.38599334335327 -149332.691102667
2.98249167919159 -147359.790211068
3.57899001502991 -147420.07217212
4.17548835086822 -145948.470004155
4.77198668670654 -145396.909030393
5.36848502254486 -142594.870439425
5.96498335838318 -142248.069729701
6.5614816942215 -138886.662535511
7.15798003005981 -136267.403096446
7.75447836589813 -135980.216826616
8.35097670173645 -136381.371812019
8.94747503757477 -134120.865108221
9.54397337341309 -133147.972581477
10.1404717092514 -132411.934212404
10.7369700450897 -129274.006918405
11.333468380928 -129627.560107793
11.9299667167664 -126802.5375764
12.5264650526047 -123232.206791655
13.122963388443 -120149.865156195
13.7194617242813 -122876.444647144
14.3159600601196 -121734.74808662
14.9124583959579 -120504.166594195
15.5089567317963 -121513.329555663
16.1054550676346 -117495.421779292
16.7019534034729 -113286.078326919
17.2984517393112 -110052.966967264
17.8949500751495 -108378.453163963
18.4914484109879 -104010.367055631
19.0879467468262 -104679.343484415
19.6844450826645 -95803.9721313451
20.2809434185028 -99606.9904784772
20.8774417543411 -95552.5325175018
21.4739400901794 -89209.9160667464
22.0704384260178 -88192.2831293613
22.6669367618561 -80943.3180577957
23.2634350976944 -78957.9644297152
23.8599334335327 -79584.2624678991
24.456431769371 -71329.6491836904
25.0529301052094 -73166.5548758529
25.6494284410477 -64512.5457378748
26.245926776886 -63373.5134129588
26.8424251127243 -57935.658068904
27.4389234485626 -56114.40838694
28.0354217844009 -55698.7680599562
28.6319201202393 -53553.7510818183
29.2284184560776 -49602.3773249973
29.8249167919159 -51520.9883146744
30.4214151277542 -49519.0779610531
31.0179134635925 -43573.8409399686
31.6144117994308 -47536.0377368673
32.2109101352692 -45735.3014765295
32.8074084711075 -38955.4065731619
33.4039068069458 -40559.0812938037
34.0004051427841 -43162.8682299197
34.5969034786224 -41036.9879424876
35.1934018144607 -35365.1284588786
35.7899001502991 -36621.7769960822
36.3863984861374 -38480.5050929705
36.9828968219757 -35001.9180421049
37.579395157814 -38688.1498792062
38.1758934936523 -37667.2268135996
38.7723918294907 -38273.7390236277
39.368890165329 -32833.0097283932
39.9653885011673 -35094.5258770653
40.5618868370056 -32936.4489570705
41.1583851728439 -33366.7213993616
41.7548835086822 -31751.8252055726
42.3513818445206 -33898.5809997893
42.9478801803589 -32899.3029028935
43.5443785161972 -38076.7143555397
44.1408768520355 -35921.4914609516
44.7373751878738 -36226.0758073754
45.3338735237122 -33965.7322897823
45.9303718595505 -32890.2409149662
46.5268701953888 -33166.8003275178
47.1233685312271 -31884.841110321
47.7198668670654 -33308.1428874684
48.3163652029037 -31980.6982733005
48.9128635387421 -37179.508528656
49.5093618745804 -32066.0392994464
50.1058602104187 -32708.4295353249
50.702358546257 -30697.3534865774
51.2988568820953 -34023.7275812511
51.8953552179336 -33336.3159286271
52.491853553772 -30820.1974754698
53.0883518896103 -33722.6911134541
53.6848502254486 -32401.9643861812
54.2813485612869 -33312.028893823
54.8778468971252 -32224.1441803205
55.4743452329636 -32954.2594816616
56.0708435688019 -31674.3758108351
56.6673419046402 -31215.2069388932
57.2638402404785 -28049.8599497777
57.8603385763168 -35292.5612100238
58.4568369121551 -29391.0999705855
59.0533352479935 -31419.4492338258
59.6498335838318 -32661.2964525168
60.2463319196701 -31246.6555271497
60.8428302555084 -32494.9447756726
61.4393285913467 -29027.6294047721
62.0358269271851 -29832.5143673244
62.6323252630234 -28735.135836853
63.2288235988617 -27742.2457051792
63.8253219347 -31042.1003645565
64.4218202705383 -29651.6467673585
65.0183186063766 -30730.4912587863
65.614816942215 -31834.3581348749
66.2113152780533 -31541.6030045086
66.8078136138916 -30954.7436680418
67.4043119497299 -27419.6902151592
68.0008102855682 -32871.2913969669
68.5973086214066 -31616.8760916992
69.1938069572449 -31755.6043431242
69.7903052930832 -31124.2411069587
70.3868036289215 -30409.368734957
70.9833019647598 -30103.9740100142
71.5798003005981 -31567.4940094548
72.1762986364365 -28790.9653076879
72.7727969722748 -29985.7804843935
73.3692953081131 -29558.2685968328
73.9657936439514 -29692.2599465223
74.5622919797897 -30967.3351709266
75.158790315628 -29527.8986762872
75.7552886514664 -27557.8732218085
76.3517869873047 -27586.9865713608
76.948285323143 -28748.1344209273
77.5447836589813 -32755.0240743357
78.1412819948196 -28664.0799989103
78.737780330658 -30194.1439875524
79.3342786664963 -30204.5677665721
79.9307770023346 -28065.316121295
80.5272753381729 -30518.2217224796
81.1237736740112 -29162.5256268474
81.7202720098495 -30095.0307037084
82.3167703456879 -25618.8541665562
82.9132686815262 -28627.3712877168
83.5097670173645 -28384.9103050979
84.1062653532028 -30478.5964486219
84.7027636890411 -24044.9433628802
85.2992620248795 -28790.495940789
85.8957603607178 -28966.0957794339
86.4922586965561 -29453.8968064068
87.0887570323944 -30274.3482549684
87.6852553682327 -29616.7844295733
88.281753704071 -26284.6039802622
88.8782520399094 -27250.3677702829
89.4747503757477 -25174.7106008446
};
\addlegendentry{$q$-SVGP}
\end{axis}

\end{tikzpicture}
   \end{subfigure}
  \hfill
  \begin{subfigure}[t]{.48\textwidth}
    \centering
\begin{tikzpicture}

\definecolor{color0}{rgb}{0.12156862745098,0.466666666666667,0.705882352941177}
\definecolor{color1}{rgb}{1,0.498039215686275,0.0549019607843137}

\begin{axis}[
height=\figureheight,
legend cell align={left},
legend style={fill opacity=0.8, draw opacity=1, text opacity=1, draw=white!80!black},
tick align=outside,
tick pos=left,
width=\figurewidth,
x grid style={white!69.0196078431373!black},
xlabel={Wall-clock time (s)},
xmin=-4.37717480368614, xmax=93.9438896700064,
xtick style={color=black},
y grid style={white!69.0196078431373!black},
ylabel={NLPD},
ymin=0.196668863922381, ymax=2.40543079631652,
ytick style={color=black}
]
\addplot [thick, color0]
table {%
0.0919644905726115 2.30503252666224
0.183928981145223 2.30191548160635
0.275893471717834 2.29551075380522
0.367857962290446 2.28826446444835
0.459822452863057 2.28007757119106
0.551786943435669 2.26658269759273
0.64375143400828 2.25388945654417
0.735715924580892 2.24139240337722
0.827680415153503 2.22453985210938
0.919644905726115 2.20967045875369
1.01160939629873 2.19172297838177
1.10357388687134 2.17585340189274
1.19553837744395 2.16194250673501
1.28750286801656 2.14678408548398
1.37946735858917 2.13153986397136
1.47143184916178 2.11837061691279
1.5633963397344 2.10380758734552
1.65536083030701 2.0906393172793
1.74732532087962 2.07511401442065
1.83928981145223 2.06007935349569
1.93125430202484 2.04243509761011
2.02321879259745 2.0246045557224
2.11518328317006 2.00143836214315
2.20714777374268 1.97794115143687
2.29911226431529 1.95313521482342
2.3910767548879 1.92609922577028
2.48304124546051 1.89445880522322
2.57500573603312 1.85811190778776
2.66697022660573 1.81818630877396
2.75893471717835 1.77589831300157
2.85089920775096 1.73016192883211
2.94286369832357 1.68135586288696
3.03482818889618 1.63069839445057
3.12679267946879 1.57519877660558
3.2187571700414 1.5164887760711
3.31072166061401 1.45953971717946
3.40268615118663 1.40029374380384
3.49465064175924 1.33820220242943
3.58661513233185 1.2764906077026
3.67857962290446 1.21644130466165
3.77054411347707 1.15453106530263
3.86250860404968 1.09309468425462
3.95447309462229 1.03095228393823
4.04643758519491 0.97255039663944
4.13840207576752 0.917019937399254
4.23036656634013 0.867939272034082
4.32233105691274 0.818939064478117
4.41429554748535 0.775187464384189
4.50626003805796 0.733280556751464
4.59822452863057 0.695835639042328
4.69018901920319 0.661858894343665
4.7821535097758 0.629855486369334
4.87411800034841 0.600512378877797
4.96608249092102 0.574205957585325
5.05804698149363 0.550621879339012
5.15001147206624 0.529502251933135
5.24197596263885 0.51161968836559
5.33394045321147 0.494689279666336
5.42590494378408 0.480831752738697
5.51786943435669 0.467924223260648
5.6098339249293 0.45726760187478
5.70179841550191 0.447018915235488
5.79376290607452 0.437803909848901
5.88572739664714 0.428783261618351
5.97769188721975 0.421906669124287
6.06965637779236 0.413250901417268
6.16162086836497 0.404611971394488
6.25358535893758 0.398654460824273
6.34554984951019 0.394257719538127
6.4375143400828 0.390217918745417
6.52947883065542 0.386532321974994
6.62144332122803 0.383052029617357
6.71340781180064 0.380265489194548
6.80537230237325 0.377137558538245
6.89733679294586 0.373646375740371
6.98930128351847 0.370872508248852
7.08126577409108 0.367533676967396
7.1732302646637 0.365196972917359
7.26519475523631 0.362914540359755
7.35715924580892 0.359754035317605
7.44912373638153 0.356353597990404
7.54108822695414 0.35294645234353
7.63305271752675 0.350828419973524
7.72501720809937 0.347174474521996
7.81698169867198 0.345512843648787
7.90894618924459 0.346228999516338
8.0009106798172 0.344378794441804
8.09287517038981 0.342932450497647
8.18483966096242 0.342511342324213
8.27680415153503 0.341796953046342
8.36876864210765 0.340651999248762
8.46073313268026 0.338339906977463
8.55269762325287 0.334912932517262
8.64466211382548 0.332787949727356
8.73662660439809 0.332547790830174
8.8285910949707 0.331529812316986
8.92055558554331 0.331122497607021
9.01252007611593 0.330085286255514
9.10448456668854 0.329471997719876
9.19644905726115 0.328515153099539
9.28841354783376 0.327351273831828
9.38037803840637 0.32958105956797
9.47234252897898 0.329872056176911
9.5643070195516 0.328569575905845
9.65627151012421 0.327052541289323
9.74823600069682 0.325034132622761
9.84020049126943 0.323233043380089
9.93216498184204 0.321570727187377
10.0241294724147 0.321624042893643
10.1160939629873 0.321203482455049
10.2080584535599 0.32012313369222
10.3000229441325 0.319254855281151
10.3919874347051 0.318693042725634
10.4839519252777 0.318622674120916
10.5759164158503 0.316843035426139
10.6678809064229 0.315895180818842
10.7598453969955 0.315552298844366
10.8518098875682 0.314005136848076
10.9437743781408 0.314181014304943
11.0357388687134 0.31310473838275
11.127703359286 0.312737178821303
11.2196678498586 0.311958399983357
11.3116323404312 0.310192200517109
11.4035968310038 0.310002444363277
11.4955613215764 0.310501664609261
11.587525812149 0.311538585032208
11.6794903027217 0.309617599919522
11.7714547932943 0.307869282346598
11.8634192838669 0.306954490910816
11.9553837744395 0.305975941184959
12.0473482650121 0.305395827248838
12.1393127555847 0.304075967221865
12.2312772461573 0.304201921180839
12.3232417367299 0.304842485335117
12.4152062273026 0.30533020648189
12.5071707178752 0.305493750091948
12.5991352084478 0.305934004497949
12.6910996990204 0.304601898943356
12.783064189593 0.302517736730384
12.8750286801656 0.303294066713889
12.9669931707382 0.303964226284613
13.0589576613108 0.304815842413098
13.1509221518834 0.304215297562248
13.2428866424561 0.303474859779201
13.3348511330287 0.304148419691246
13.4268156236013 0.304223659940054
13.5187801141739 0.303887379134386
13.6107446047465 0.301811464075295
13.7027090953191 0.299615929412295
13.7946735858917 0.297814296899824
};
\addlegendentry{$t$-SVGP}
\addplot [thick, color1]
table {%
0.596498335838318 2.30345680493316
1.19299667167664 2.30047419335222
1.78949500751495 2.29195630862771
2.38599334335327 2.28283439492993
2.98249167919159 2.2729052221698
3.57899001502991 2.26085196497334
4.17548835086822 2.24583728359946
4.77198668670654 2.23090525486484
5.36848502254486 2.21771451085353
5.96498335838318 2.20426499407019
6.5614816942215 2.18482625042799
7.15798003005981 2.16787794840032
7.75447836589813 2.15145226972992
8.35097670173645 2.13895451175539
8.94747503757477 2.12323425434017
9.54397337341309 2.10947977427971
10.1404717092514 2.09390461794579
10.7369700450897 2.07754722683287
11.333468380928 2.06329495910171
11.9299667167664 2.04623380974686
12.5264650526047 2.02745519408482
13.122963388443 2.00877508976478
13.7194617242813 1.98538722709463
14.3159600601196 1.96091793975596
14.9124583959579 1.93129690266641
15.5089567317963 1.89936832999721
16.1054550676346 1.86329751622704
16.7019534034729 1.82433626560169
17.2984517393112 1.77989434458312
17.8949500751495 1.73120112158445
18.4914484109879 1.68079201290935
19.0879467468262 1.62620999619681
19.6844450826645 1.56661764407055
20.2809434185028 1.505874941174
20.8774417543411 1.44078535154997
21.4739400901794 1.3755456869491
22.0704384260178 1.30653752642294
22.6669367618561 1.23690880852524
23.2634350976944 1.16775886172599
23.8599334335327 1.09791277257676
24.456431769371 1.02829510005485
25.0529301052094 0.963970023428875
25.6494284410477 0.902420979986043
26.245926776886 0.844011125846981
26.8424251127243 0.789781580010316
27.4389234485626 0.740904126078144
28.0354217844009 0.694737108149315
28.6319201202393 0.654004544857632
29.2284184560776 0.618589766309734
29.8249167919159 0.586759554138895
30.4214151277542 0.557970815755297
31.0179134635925 0.532173209592955
31.6144117994308 0.509750229011086
32.2109101352692 0.490014466072351
32.8074084711075 0.472973005337144
33.4039068069458 0.457987670341672
34.0004051427841 0.444302351158948
34.5969034786224 0.431847829402973
35.1934018144607 0.42286905527934
35.7899001502991 0.414333936511095
36.3863984861374 0.408990941911692
36.9828968219757 0.404474079742234
37.579395157814 0.396704285314907
38.1758934936523 0.389660823205763
38.7723918294907 0.38443420528154
39.368890165329 0.379670507681
39.9653885011673 0.372914937034302
40.5618868370056 0.368238618294623
41.1583851728439 0.363801978488469
41.7548835086822 0.360229141735789
42.3513818445206 0.359082178273366
42.9478801803589 0.356620046904691
43.5443785161972 0.353070679384644
44.1408768520355 0.349862280904349
44.7373751878738 0.347586679615285
45.3338735237122 0.346327038493899
45.9303718595505 0.345792981562949
46.5268701953888 0.344922285668009
47.1233685312271 0.343820089243199
47.7198668670654 0.342790630734895
48.3163652029037 0.340760342505067
48.9128635387421 0.338771012652986
49.5093618745804 0.336878649534214
50.1058602104187 0.335867401344538
50.702358546257 0.335346272504609
51.2988568820953 0.334977230546834
51.8953552179336 0.333756786156695
52.491853553772 0.33415729151278
53.0883518896103 0.331662559412035
53.6848502254486 0.330137958391062
54.2813485612869 0.326660736477431
54.8778468971252 0.327157118626989
55.4743452329636 0.329203900978903
56.0708435688019 0.329367128172084
56.6673419046402 0.327030229387144
57.2638402404785 0.325822655704393
57.8603385763168 0.323316640945901
58.4568369121551 0.32215593105681
59.0533352479935 0.321409596947766
59.6498335838318 0.323385721616683
60.2463319196701 0.322101216058978
60.8428302555084 0.317078567842594
61.4393285913467 0.314330329958143
62.0358269271851 0.314605120681435
62.6323252630234 0.314691062622865
63.2288235988617 0.312073008327351
63.8253219347 0.310550828727829
64.4218202705383 0.308637478423124
65.0183186063766 0.308280580046476
65.614816942215 0.307950045731565
66.2113152780533 0.30830372286994
66.8078136138916 0.308828727390149
67.4043119497299 0.310404059967827
68.0008102855682 0.309713645536493
68.5973086214066 0.3101844561526
69.1938069572449 0.310768246859334
69.7903052930832 0.312379091255196
70.3868036289215 0.30939593170466
70.9833019647598 0.308233490070792
71.5798003005981 0.309573288996859
72.1762986364365 0.30893175923354
72.7727969722748 0.306827263854001
73.3692953081131 0.305192248525793
73.9657936439514 0.305172083594016
74.5622919797897 0.303826664782256
75.158790315628 0.301757966091933
75.7552886514664 0.303495455279541
76.3517869873047 0.304996097876643
76.948285323143 0.304062787091775
77.5447836589813 0.301771953202389
78.1412819948196 0.299380507465254
78.737780330658 0.29966319403716
79.3342786664963 0.298916331513115
79.9307770023346 0.300543449185084
80.5272753381729 0.302563161961264
81.1237736740112 0.304285338337269
81.7202720098495 0.301499042315008
82.3167703456879 0.301483309022174
82.9132686815262 0.301989201969854
83.5097670173645 0.306370369370544
84.1062653532028 0.304998896347868
84.7027636890411 0.306104708466823
85.2992620248795 0.306507236567406
85.8957603607178 0.302445564537394
86.4922586965561 0.302037292312994
87.0887570323944 0.303234775884349
87.6852553682327 0.303170796184067
88.281753704071 0.300030015316065
88.8782520399094 0.297744540692983
89.4747503757477 0.29706713357666
};
\addlegendentry{$q$-SVGP}
\end{axis}

\end{tikzpicture}
   \end{subfigure}\\[-3pt]
  \caption{Comparison of practical inference and learning on MNIST. We compare training time on a laptop between $t$-SVGP to the $q$-SVGP model in GPflow in terms of wall-clock time of training for 150 steps, where both methods use natural gradient updates and share the same learning rates.}
  \label{fig:mnist}
\end{figure}
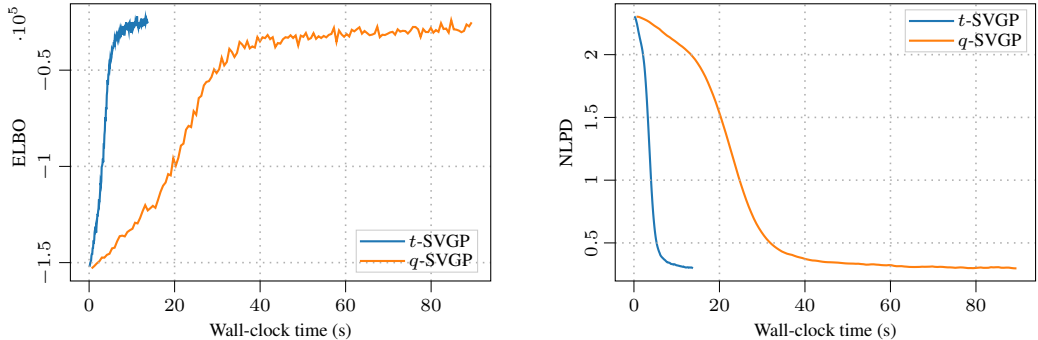

\section{Discussion and Conclusion}
\label{sec:conclusion}
Sparse variational GP (SVGP) methods are the current {\it de~facto} approach to allow GPs to scale to large problems. In this paper, we introduced an alternative parameterization to variational GPs
that leads to an improved loss landscape for learning (\cf, \cref{fig:bound}). This improvement hinges on writing the variational problem in terms of its dual---similar to the conjugate-computation variational inference (CVI) approach by \citet{khan2017conjugate}---parameterization to capture sites: we assume the approximate posterior decomposes into a prior contribution and a Gaussian {\em approximate} likelihood contribution. Variational inference under this model can conveniently be implemented by mirror descent and corresponds to natural gradient based learning, thus improving convergence
in variational parameter optimization 
(the `E-step'), at the same time as improving hyperparameter optimization (the `M-step'), due to the tighter evidence lower bound.

We further show that we can derive the {\em sparse} equivalent of this method, which also allows for stochastic training through mini-batching, reducing the computational complexity to $\order(m^3 + n_\mathrm{b} m^2)$ per step. Our method matches the asymptotic computational cost of other SVGP methods, while marginally reducing compute due to simpler expressions to back-propagate through (see discussion in \cref{sec:experiments}). 
Our empirical validation across a wide variety of regression and classification tasks confirms the benefits suggested by our theory: The proposed strategy typically allows for improved stability over gold-standard SVGP methods even when the learning rates remain the same. It allows for higher learning rates, and reduces computational cost---leading to improved learning both in terms of reduced steps as well as expected wall-clock time.

We provide a reference implementation of our method under the GPflow framework at \url{https://github.com/AaltoML/t-SVGP}.

\begin{ack}

AS acknowledges funding from the Academy of Finland (grant numbers 324345 and 339730). We acknowledge the computational resources provided by the Aalto Science-IT project. 
We thank Stefanos Eleftheriadis, Richard E.\ Turner, and Hugh Salimbeni for comments on the manuscript.

\end{ack}

\phantomsection%
\addcontentsline{toc}{section}{References}
\begingroup
\small
\bibliographystyle{abbrvnat}

\endgroup

\endgroup

\clearpage

\appendix

\nipstitle{{\Large Supplementary Material for} \\
  Dual Parameterization of Sparse Variational Gaussian Processes}

\section{Tighter Bound for the M-step}
\label{app:tighter_bound}

We here study the role of parameterizations $\vxi$ in shaping the losses optimized during the M-step of the EM learning procedure. Each parameterization $\vxi$ has associated natural parameters $\veta$.

We introduce an alternative expression  of the loss $\cL$ in terms of the natural parameters of the prior $\veta_p$ and of the approximate posterior $\veta_q$: $\cL(q, \vtheta) = L(\veta_q,\veta_p)$. To simplify the presentation but without loss of generality, we consider the case $\vtheta=\veta_p$, i.e. when the hyperparameters are directly the natural parameters. The case we actually care about is when $\vtheta$ indexes natural parameters $\veta_p(\vtheta)$, in which case, the natural parameters lie on a manifold in $\Omega$.

We focus on the difference between parameterizations where the posterior  statistics $\veta_q$ depends on the prior statistics $\veta_p$, as in the dual parameterization $\vlambda$, where this dependence is linear $\veta_q = \veta_p + \vlambda$, versus parameterizations that don't, as in the $\vxi=(\vmu, \ML)$ parameterization.
To make this distinction explicit we introduce the losses 
\begin{align}
    \tilde{l}(\veta_p) &= L(\veta_p + \vlambda^*,\veta_p) , \\
    l(\veta_p) &= L(\veta_q^*,\veta_p) .
\end{align}

For a matched optimal E-step, \ie\ ${\veta_p+\vlambda^*=\veta_q^* = \argmax_{\veta}L(\veta,\veta_p)}$, the value of $l$ and $\tilde{l}$ and their gradient w.r.t. $\veta_p$ are the same:
\begin{align}
\tilde{l}(\veta_p) &=  l(\veta_p) , \\
\nabla_{\veta_p} \tilde{l}(\veta_p)  &= \underbrace{\partial_{\veta_1} L |_{\veta_{q^*}}}_{=0} + \partial_{\veta_2} L |_{\veta_{p}}  = \partial_{\veta_2} \cL |_{\veta_{p}} = \nabla_{\veta_p} l(\veta_p) .
\end{align}

In the conjugate regression case, we have that $\tilde{l}(\veta_p) \geq l(\veta_p)$:
\begin{align}
\tilde{l}(\veta_p) - l(\veta_p) &=
- (\log p(\data) -\tilde{l}(\veta_p)) + (\log p(\data) - l(\veta_p)) \\
&= -\underbrace{\KL{\veta_p + \vlambda^*}{\veta_{\text{post}}}}_{=0} + \KL{\veta_{q^*}}{\veta_{\text{post}}}\\
&=  \KL{\veta_{q^*}}{\veta_{\text{post}}} > 0 .
\end{align}
We can't show this in the non-conjugate setting but instead focus on the local behavior of $\tilde{l}(\veta_p)$ and $l(\veta_p)$. Specifically, since their gradients match, we study their Hessians, which are different:
\begin{align}
  \nabla^2_{\veta_p\veta_p}\,\tilde{l}(\veta_p)
 &= \partial^2_{\veta_1\veta_1} L|_{\veta_{q^*}} + \partial^2_{\veta_2\veta_2} L|_{\veta_{p}}
 +2\partial^2_{\veta_1\veta_2} L|_{\veta_{q^*}\veta_{p}}\\
   \nabla^2_{\veta_p\veta_p}\,l(\veta_p)   
 &=  \partial^2_{\veta_2\veta_2} L|_{\veta_{p}}
 \end{align}

The Hessian difference between the two conditions is 
\begin{align}
\Delta H &=  \partial^2_{\veta_1\veta_1} L|_{\veta_{q^*}} +2\partial^2_{\veta_1\veta_2} L|_{\veta_{q^*}\veta_{p}}
 \end{align}

and using the identity
\begin{align}
\partial^2_{\veta_1\veta_2} L|_{\veta_{q^*}\veta_{p}} &= -
\partial^2_{\veta_1\veta_2} D_{KL}(\veta_{t}+\veta_{p},\veta_{p})|_{\veta_{q^*}\veta_{p}}
 = \MI[\veta_{q^*}] .
 \end{align}
The Hessian difference can be expressed as 
\begin{align}
\Delta H &= \partial^2_{\veta_1\veta_1} L|_{\veta_{q^*}}+2\MI[\veta_{q^*}] .%
 \end{align}

$\tilde{l}(\veta_p)$ is a local upper bound to $l(\veta_p)$ if $\Delta H \succeq 0$
\begin{equation}
\Delta H \succeq 0 	
\iff
\partial^2_{\veta_1\veta_1} L|_{\veta_{q^*}}
 \succeq -  2\MI[\veta_{q^*}] .
\end{equation}
This corresponds to a condition on the curvature of the optimization problem in the preceding E-step. We can verify that this condition is met in the conjugate case where 
\begin{equation}
    \partial^2_{\veta_1\veta_1} L|_{\veta_{q^*}} =  - \partial^2_{\veta_1\veta_1} \KL{\veta_{q^*}}{\veta_{\text{post}}} = - \MI[\veta_{q^*}].
\end{equation}
The condition is indeed met since the Fisher information matrix $\MI[\veta_{q^*}]$ is positive semi-definite.

\section{Proposed Objective for the M-step of $t$-SVGP}
\label{app:proposed_elbo}

Starting from hyperparameter $\vtheta_{\text{old}}$, an E-step gives the optimal dual parameters $\vlambda^*$. The objective for the proposed M-step of $t$-SVGP is the ELBO in \cref{eq:elbo_svgp_new} for the variational distribution $q_{\vu}(\vu; \hat{\veta}_{\vu}(\vtheta))$ with $\vtheta$ dependent parameters $\hat{\veta}_{\vu}(\vtheta)$ expressed in terms of the  mean and covariance matrix as
\begin{equation}
     \hat{\MS}_{\vu}\inv\hat{\vm}_{\vu} = \MKuu^{-1} \underbrace{ \rnd{ \msum_{i=1}^{n} \vk_{\vu i} {\lambda}_{1,i}^*} }_{= \bar{\vlambda}_1}
     \text{ and }
     \hat{\MS}_{\vu}\inv = \MKuu\inv + \MKuu\inv \underbrace{ \rnd{ \msum_{i=1}^{n} \vk_{\vu i} {\lambda}_{2,i}^* \vk_{\vu, i}^\top }}_{= \bar{\MLambda}_2} \MKuu\inv.
\end{equation}
Introducing ${q}(\vf, \vu; \vtheta)=p_{\vtheta}(\vf| \vu) q_{\vu}(\vu; \hat{\veta}_{\vu}(\vtheta))$, 
the ELBO for our proposed M-step is given by:
\begin{align}
    \cL_{\eta_{u}}(\hat{\veta}_{\vu}(\vtheta), \vtheta) 
    &= \myexpect_{{q}(\vf, \vu; \vtheta)}  \sqr{ \log \frac{p_{\theta}(\vy, \vf, \vu)}{\hat{q}_t(\vf, \vu; \vtheta)} } \nonumber \\
    &= \myexpect_{{q}(\vf, \vu; \vtheta)} \sqr{ \log \frac{\prod_{i=1}^n p(y_i \mid f_i) \cancel{ p_{\vtheta}(\vf \mid \vu)p_{\vtheta}(\vu)} }{ \frac{1}{\mathcal{Z}(\vtheta)}  t^*(\vu) \cancel{p_{\vtheta}(\vf \mid \vu)p_{\vtheta}(\vu)} } } \nonumber \\
    &=  \log \mathcal{Z}(\vtheta) + c(\vtheta), \vphantom{\bigg|}
\end{align}
where 
$c(\vtheta)=\msum_{i=1}^n \myexpect_{{q}_t(f_i; \vtheta)}[\log p(y_i \mid f_i)] - \myexpect_{{q}_t(\vu; \vtheta)}[{\log\,t^*(\vu)}]$ and
$\log \mathcal{Z}(\vtheta)$ is the log-partition of the Gaussian $ q_{\vu}(\vu; \hat{\veta}_{\vu}(\vtheta))$ 
\begin{multline}
        \log \mathcal{Z}(\vtheta) = -\frac{m}{2} \log (2\pi) - \frac{1}{2} \log| \MKuu(\vtheta)\bar{\MLambda}_2\inv\MKuu(\vtheta) + \MK_{\vu\vu}(\vtheta)| \\ - \frac{1}{2} \widetilde{\vy}^\top \sqr{ \MKuu(\vtheta)\bar{\MLambda}_2\inv\MKuu(\vtheta) + \MK_{\vu\vu}(\vtheta) }^{-1} \widetilde{\vy},
        \label{eq:Zt_SVGP}
\end{multline}
with $\widetilde{\vy} = \MKuu(\vtheta)\bar{\MLambda}_2\inv
\bar{\vlambda}_1$.

\section{Efficient ELBO Computation for $t$-SVGP}
\label{sec:computation}
We here detail the computations required to perform inference and learning using the dual parameterization.
To perform inference, 
the variational expectations need to be evaluated. These require the evaluation of the marginal predictions $q(f(\vx_i))$ for all inputs $\vx_i$ in $\data$.
For learning, the ELBO in \cref{eq:elbo_svgp_new} needs to be evaluated which requires the computation of a KL divergence.

In $t$-SVGP, the variational distribution $q(\vu)= \N(\vu| \vm, \MS)$ is parameterized in terms of its natural parameters:
\begin{align} 
\MS^{-1} &= \MKuu^{-1} + \MKuu^{-1}\bar{\MLambda}_2\MKuu^{-1} , \\
\MS^{-1}\vm &= \MKuu^{-1}\bar{\vlambda}_1,
\end{align}
where
\begin{equation}
\label{eq:lambda_mxm2}
     \bar{\vlambda}_1 =  \msum_{i=1}^{n} \vk_{i\vu}^\top {\lambda}_{1,i}
     \qquad \text{and} \qquad
     \bar{\MLambda}_2 =  \msum_{i=1}^{n}\vk_{i\vu}^\top \vk_{i\vu} {\lambda}_{2,i} .
\end{equation}

Introducing $\MR=\MKuu +\bar{\MLambda}_2$, the mean and covariance $q(\vu)$ can be rewritten as:
\begin{align} 
\MS &= (\MKuu^{-1} +  \MKuu^{-1}\bar{\MLambda}_2\MKuu^{-1})^{-1}\\
 &= \MKuu(\MKuu +\bar{\MLambda}_2)^{-1}\MKuu\\
 &=\MKuu \MR^{-1} \MKuu,
 \\
\vm &= \MKuu \MR^{-1} \bar{\vlambda}_1.
\end{align}
This leads to simple closed form expressions for the marginal predictions:
\begin{equation} 
q(\vf^\star)
= \N(\vf^\star | \MKsu \MR^{-1} \bar{\vlambda}_1, \MKss - \MKsu\MKuu^{-1}\MKus + \MKsu\MR^{-1}\MKus),
\end{equation}
and for the and KL divergence \cref{eq:elbo_svgp_new}:
\begin{align}
  \mathrm{D}_\text{KL}\left(q(\vu) \parallel p(\vu)\right) &=
  \tfrac{1}{2}\left(
    \tr\left(\MKuu^{-1}\MS\right) +
    \vm^{\top} \MKuu^{-1}\vm - k +
    \ln|\MKuu\MS\inv|
  \right) \\
 &=
 \tfrac{1}{2}\left(
       \tr(\MKuu\MR^{-1}  ) - k +\vlambda_1^{\top}\MR^{-1} \MKuu\MR^{-1} \bar{\vlambda}_1
       -\ln |\MKuu| + \ln |\MR|
        \right).
\end{align}

\begin{comment}

%
We introduce $\MR=\MKuu +\MLambda_2 $, via
%
\begin{align} 
\MLambda^{-1} &= (\MKuu^{-1} +  \MKuu^{-1}\MLambda_2\MKuu^{-1})^{-1}\\
 &= \MKuu(\MKuu +\MLambda_2)^{-1}\MKuu\\
 &=\MKuu \MR^{-1} \MKuu
 \\
\vm &= \MKuu \MR^{-1} \MLambda_1
\end{align}
%
Posterior predictions are given by
%
\begin{align} q(\vf^\star) &= \mathcal N(\vf^\star| \MKsu \MR^{-1}\MLambda_1, \MKss - \MKsu\MKuu^{-1}\MKus +\MKsu\MR^{-1}\MKus)\end{align}
%
And the Kullback-Leibler divergence by
%
\begin{align}
  \mathrm{D}_\text{KL}\left(N_q \parallel N_p\right) &=
  \tfrac{1}{2}\left(
    \tr\left(\MKuu^{-1}\MLambda^{-1}\right) +
    \mu_q^{\top} \MKuu^{-1}\mu_q - k +
    \ln|\MKuu\MLambda|
  \right).\\
 &=
 \tfrac{1}{2}\left(
       \tr(\MKuu\MR^{-1}  ) - k +\vlambda^{(1)\top}\MR^{-1} \MKuu\MR^{-1} \MLambda_1
       -\ln |\MKuu| + \ln |\MR|
        \right).
\end{align}

\end{comment}

%
\label{sec:datasets}

\section{Pseudocode for the $q$-SVGP Algorithm} 
\label{app:pseudo-q-SVGP}
We here detail the $q$-SVGP algorithm for inference and learning with the E-step as described in \cite{salimbeni2018natural}, for parameterization $\vxi = (\vm, \ML)$.
The pseudocode shows an E-step comprised of $K$ iterations of natural gradient descent, followed by an M-step comprised of $S$ gradient descent iterations with learning rate $\gamma$. 
\begin{algorithm}[H]
\caption{$q$-SVGP}
\renewcommand{\algorithmiccomment}[1]{\bgroup\hfill{\footnotesize \it \color{gray}#1}\egroup}
\algsetup{linenosize=\footnotesize}
\label{alg:natgrads} 
\begin{algorithmic}[1]
\STATE initialization at $\vtheta_t$, $\vxi_t$
 \FOR {$k = 0\dots K-1$} 
\STATE $\vxi^{(0)}\gets \vxi_t$ \COMMENT{Initialization of the natural gradient descent iterations}
  \FOR {$i = 1\dots n$} 
\STATE $q^{(k)}(f_i) = \int p_{\vtheta_t}(f_i \mid \vu)q_{\vu}^{(k)}(\vu)\,\dee\vu$  \COMMENT{Marginal predictions}
\ENDFOR
\STATE $\cL^{(k)} = \sum_i\Exp{q^{(k)}(f_i)}{\log p(y_i\mid f_i)} - \KL{q_{\vu}^{(k)}(\vu)}{p_{\vtheta_t}(\vu)}$ \COMMENT{ELBO}
\STATE $\veta^{(k)} \gets \vxi^{(k)} $ \COMMENT{Gaussian transformation}
\STATE $\vg^{(k)} \gets 
 \nabla_{\vmu}\vxi(\vmu^{(k)})  \nabla_{\vxi}\cL^{(k)}|_{\vxi=\vxi_t}$ \COMMENT{Natural gradient}
 \STATE $\veta^{(k+1)} \gets \veta^{(k)} + \rho \,
\vg^{(k)}$ \COMMENT{Natural gradient step}
\STATE $\vxi^{(k+1)} \gets \veta^{(k+1)}$ \COMMENT{Gaussian transformation}
\ENDFOR
\STATE $\vxi_{t+1} \gets \vxi^{(K)}$ \COMMENT{End of E-step}
\STATE $\vtheta^{(0)}\gets \vtheta_t$ \COMMENT{Initialization of the gradient descent iterations}
 \FOR {$s = 0\dots S-1$} 
\STATE $\tilde{\cL}^{(s)}(\vtheta) = -\KL{q^{(s)}_\vu(\vu)}{p_{\vtheta}(\vu)}$ \COMMENT{KL of ELBO}
\STATE $\vtheta^{(s+1)} \gets \vtheta^{(s)} + \gamma \nabla_{\vtheta}\tilde{\cL}^{(s)}|_{\vtheta=\vtheta^{(s)}}$ \COMMENT{Gradient descent step for $\vtheta$}
\ENDFOR
\STATE $\vtheta_{t+1} \gets \vtheta^{(S)}$ \COMMENT{End of M-step}
\end{algorithmic}
\end{algorithm}

\section{Pseudocode for the $t$-SVGP Algorithm} 
\label{app:pseudo-t-SVGP}
We here summarize the $t$-SVGP algorithm using the dual parameterization.
The pseudocode shows an E-step comprised of $K$ iterations of natural gradient descent, followed by an M-step comprised of $S$ gradient descent iterations with learning rate $\gamma$.

\begin{algorithm}[H]
\caption{$t$-SVGP}
\renewcommand{\algorithmiccomment}[1]{\bgroup\hfill{\footnotesize \it \color{gray}#1}\egroup}
\algsetup{linenosize=\footnotesize}
\label{alg:natgrads} 
\begin{algorithmic}[1]
\STATE initialization at $\vtheta_t$, $\vlambda_t$
 \FOR {$k = 0\dots K-1$} 
 \STATE $\vlambda^{(0)}\gets \vlambda_t$ \COMMENT{Initialization of the natural gradient descent iterations}
 \FOR {$i = 1\dots n$} 
\STATE $q^{(k)}_{\vu}(f_i) = \int p_{\vtheta_t}(f_i \mid \vu)q_{\vu}^{(k)}(\vu; \vlambda^{(k)})\,\dee\vu$ \COMMENT{Marginal predictions}
\STATE $\alpha_i^{(k)} = \myexpect_{q_{\vu}^{(k)}(f_i)} \sqr{  \nabla_{f} \log p(y_i \mid f_i) }$
\STATE $\beta_i^{(k)} = \myexpect_{q_{\vu}^{(k)}(f_i)} \sqr{ -\nabla_{ff}^2 \log p(y_i \mid f_i) } $
\STATE $\vg_i^{(k)} = (\beta_i^{(k)} m_i^{(k)} + \alpha_i^{(k)}, \,\, \beta_i^{(k)})$ \COMMENT{Natural gradient}
\ENDFOR
\STATE $\bar{\vlambda}_1^{(k+1)} \leftarrow (1-r) \bar{\vlambda}_1^{(k)} + r \msum_{i \in \mathcal{M}} \vk_{\vu i} \vg_{1,i}^{(k)}$\COMMENT{Natural gradient step}
\STATE $\bar{\MLambda}_2^{(k+1)} \leftarrow (1-r) \bar{\MLambda}_2^{(k)} + \, r \msum_{i \in \mathcal{M}} \vk_{\vu i}\vk_{\vu i}^\top \vg_{2,i}^{(k)}$\COMMENT{Natural gradient step}
\ENDFOR
\STATE $\vlambda_{t+1} \gets \vlambda^{(K)}$ \COMMENT{End of E-step}
\STATE $\vtheta^{(0)}\gets \vtheta_t$ \COMMENT{Initialization of the gradient descent iterations}
 \FOR {$s = 0\dots S-1$} 
\STATE $\tilde{\cL}^{(s)}(\vtheta) = \log \mathcal{Z}^{(s)}(\vtheta) + c^{(s)}(\vtheta)$  \COMMENT{ELBO}
\STATE $\vtheta^{(s+1)} \gets \vtheta^{(s)} + \gamma \nabla_{\vtheta}\tilde{\cL}^{(s)}(\vtheta)$ \COMMENT{Gradient step for $\vtheta$}
\ENDFOR
\STATE $\vtheta_{t+1} \gets \vtheta^{(S)}$ \COMMENT{End of M-step}
\end{algorithmic}
\end{algorithm}

\section{Data Sets and Experimental Details}
\label{sec:datasets}

\subsection{UCI Data Sets}
For the regression experiments, we ran the E-step with a learning rate of $1$. The update amounts to a closed form GP regression step given we have a conjugate model. We then ran the M-step 15 iterations with a learning rate of $0.2$.  In the classification examples we do not have closed form updates and so ran the E-step $8$ times with a learning rate of $0.7$.  The M-step was ran the same way as in regression experiments. All other specifications where the same in all experiments. We choose $m=50$ and given the data sizes were small, we set the mini batch to equal the data size $m_b = n$, so non stochastic gradients. The inducing points were initialized by K-means and optimized in the M-step along with hyper parameters. We ran all experiments a total of $20$ full EM iterations. We ran 5-fold cross validation and in \cref{fig:uci} plotted the mean result of the the folds. The kernel used was a Mat\'ern-$5/2$ with lengthscale and amplitude both initialised at $1$ similarly if a Gaussian likelihood was used it was likewise initialised to $1$.  We now detail each data set:
\textbf{Airfoil}: The airfoil self-noise data set is regression task to predict scaled sound pressure. The data set has $d=5$ and $n=1503$ entries.
\textbf{Boston housing}: The task is to predict the median value of owner-occupied homes. The data set has $d=12$ and $n=506$ entries. 
\textbf{Concrete}: The concrete compression data set is another regression experiment, where the goal is predict concrete compressive strength with $d=5$ and $n=1030$.
\textbf{Sonar}: The data set is a classification example so we use a binomial likelihood. The goal is to predict from some sonar information if an object is a rock or a mine, the number of features is $d=60$ and number of data points $n=208$.
\textbf{Ionosphere}: Another classification example where, `Good' radar shows evidence of some type of structure in the ionosphere and "Bad" no evidence. The ionosphere data set has $n=351$ and $d=34$.
\textbf{Diabetes}: The goal of the diabetes experiment is based on patient medical information can we predict the diabetic outcome. The data consists of $d=8$ and $n=768$ entries. 

\subsection{MNIST Experiments} \label{sec:mnist}
MNIST \cite{lecun1998mnist}, available under CC BY-SA 3.0, is a handwritten digit classification task for digits 0--9. We used a softmax likelihood with 10 latent GPs, one for each digit. The data set is $n=70,000$ and $d=256$. We again used a Mat\'ern-$5/2$ covariance function and set the number of inducing points $m=100$ and used a minibatch size of $n_\mathrm{b} = 200$. The kernel lengthscale $\ell$ and amplitude $\sigma^2$ were both initialised to $1$ and the inducing points were randomly initialised. We alternated between different learning rates and number of E and M-steps as detailed in \cref{tbl:comparison}.

\subsection{Illustrative Examples}
For \cref{fig:bound} (right) and \cref{fig:fixed_point_classif} the experimental set up was similar.
We considered a simplified one-dimensional GP classification task simulated by thresholding a noisy sinc function and simulating $n=100$ observations. We considered $m=10$ equally spaced inducing points for this task and fixed the lengthscale hyperparameter to $\ell = \nicefrac{1}{2}$.

\subsection{Additional Experiments}
\begin{figure}[t]
  \centering\scriptsize
  \setlength{\figurewidth}{.4\textwidth}
  \setlength{\figureheight}{.66\figurewidth}
  \pgfplotsset{axis on top,scale only axis,y tick label style={rotate=90},grid=major, legend style={fill=white}, every y tick scale label/.style={at={(rel axis cs:-0.1,.85)},anchor=south west,inner sep=1pt}}
  \begin{subfigure}[t]{.48\textwidth}
    \centering
\begin{tikzpicture}

\definecolor{color0}{rgb}{0.12156862745098,0.466666666666667,0.705882352941177}
\definecolor{color1}{rgb}{1,0.498039215686275,0.0549019607843137}

\begin{axis}[
height=\figureheight,
legend cell align={left},
legend style={
  fill opacity=0.8,
  draw opacity=1,
  text opacity=1,
  at={(0.03,0.97)},
  anchor=north west,
  draw=white!80!black
},
tick align=outside,
tick pos=left,
width=\figurewidth,
x grid style={white!69.0196078431373!black},
xlabel={Number of inducing points, $m$},
xmin=40, xmax=260,
xtick style={color=black},
y grid style={white!69.0196078431373!black},
ylabel={Wall-clock speed (sec) per 100 iterations},
ymin=1.21814220547676, ymax=124.039681380987,
ytick style={color=black}
]
\addplot [semithick, color0, mark=x, mark size=3, mark options={solid}]
table {%
50 6.80093944072723
100 12.9318760633469
150 22.8484766483307
200 29.5031651258469
250 43.0016676187515
};
\addlegendentry{$t$-SVGP}
\addplot [semithick, color1, mark=*, mark size=2.1, mark options={solid}]
table {%
50 87.742732167244
100 90.2742160558701
150 102.566576957703
200 103.12547659874
250 118.456884145737
};
\addlegendentry{$q$-SVGP}
\end{axis}

\end{tikzpicture}
   \end{subfigure}
  \hfill
\\[-3pt]
  \caption{Wall-clock speed for $q$-SVGP and $t$-SVGP as a function of the number of inducing points $m$ on the MNIST experiment.}
  \label{fig:m_change}
\end{figure}
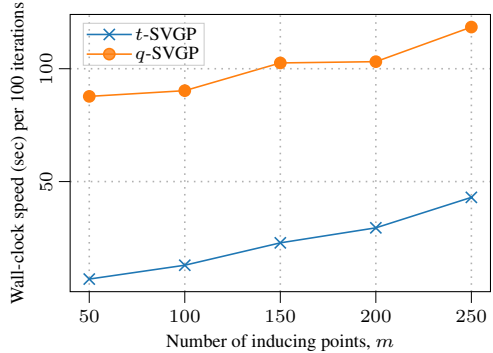

We include \cref{fig:m_change} to show the effect of changing the number of inducing points on the wall-clock speed. The experiment is the same as in \cref{sec:mnist} but we now run only for 100 iterations of a single E and M step. The chart shows that there is a constant factor caused by our computationally cheaper E-step, the effect is substantial in most practical settings where $m$ is set below 250. %

\section{Author Contributions}
The idea of dual parameterization presented in the first part of Sec.~3 and the new lower bound discussed in Sec.~3.1 is due to MEK. The idea of using the dual parameterization to speed up SVGP was conceived by PEC and VA, who derived the bound, with inspiration from separate prior work by PEC, VA, and AS. PEC had the main responsibility of implementing the methods and conducting the experiments, and VA of formalizing the methods. All authors contributed to finalizing the manuscript.

\phantomsection%
\addcontentsline{toc}{section}{References}
\begingroup
\small
\bibliographystyle{abbrvnat}

\endgroup

\end{document}